\pdfoutput=1
\documentclass{article} 
\usepackage{longcat_report,times}
\usepackage[table]{xcolor}

\usepackage{amsmath,amsfonts,bm}

\def\eqref#1{equation~\ref{#1}}

\def\1{\bm{1}}

\DeclareMathAlphabet{\mathsfit}{\encodingdefault}{\sfdefault}{m}{sl}
\SetMathAlphabet{\mathsfit}{bold}{\encodingdefault}{\sfdefault}{bx}{n}

\usepackage{enumitem}
\usepackage{hyperref}
\usepackage{url}
\usepackage[utf8]{inputenc} 
\usepackage[T1]{fontenc}    
\usepackage{hyperref}       
\usepackage{url}            
\usepackage{booktabs}       
\usepackage{amsfonts}       
\usepackage{nicefrac}       
\usepackage{wrapfig}
\usepackage{caption}
\usepackage{wrapfig}
\usepackage{graphicx}
\usepackage{microtype}      
\usepackage{xcolor}         
\usepackage{graphicx}
\usepackage{subfigure}
\usepackage{mathtools}
\usepackage{amsthm}
\usepackage{amsmath}
\usepackage{amssymb}
\usepackage{extarrows}
\usepackage{diagbox}
\usepackage{microtype}
\usepackage{multirow}
\usepackage[most]{tcolorbox}
\usepackage{xcolor}
\usepackage{booktabs}
\usepackage{tabularx}
\definecolor{systemblue}{RGB}{235,243,252}
\definecolor{usergray}{RGB}{245,245,245}
\definecolor{assistantgreen}{RGB}{237,248,240}

\newtcolorbox{systemmessage}{
  colback=systemblue,
  colframe=blue!45,
  boxrule=0.5pt,
  arc=2pt,
  left=3pt,right=3pt,top=2pt,bottom=2pt
}
\newtcolorbox{usermessage}{
  colback=usergray,
  colframe=black!25,
  boxrule=0.5pt,
  arc=2pt,
  left=3pt,right=3pt,top=2pt,bottom=2pt
}
\newtcolorbox{assistantmessage}{
  colback=assistantgreen,
  colframe=green!45!black,
  boxrule=0.5pt,
  arc=2pt,
  left=3pt,right=3pt,top=2pt,bottom=2pt
}
\usepackage{makecell}

\newcommand{\method}{\texttt{KPHarness}}

\definecolor{text}{rgb}{0.4,0.1,0.4}

\usepackage{algorithm}
\usepackage{algorithmic}

\title{KPHarness: Co-Evolving Knowledge and Probing for Long-Horizon Gaming Harnesses}

\author{
Zhikang Chen\textsuperscript{1}\quad
Yueqing Sun\textsuperscript{2}\quad
Sen Cui\textsuperscript{3}\quad
Ziang Ye\textsuperscript{2}\quad
Yu Wang\textsuperscript{2}\quad\\[3pt]
Yan Feng\textsuperscript{2,\dag}\quad
Xierui Song\textsuperscript{2,\dag}\quad
Yu Zhang\textsuperscript{4,\dag}\quad
Tingting Zhu\textsuperscript{1,\dag}\\[6pt]
{\footnotesize \textsuperscript{1}University of Oxford\quad
\textsuperscript{2}Meituan\quad
\textsuperscript{3}Tsinghua University}\\[2pt]
{\footnotesize \textsuperscript{4}Southern University of Science and Technology}\\[5pt]
{\scriptsize
\textsuperscript{\dag}Corresponding authors.}
}

\begin{document}

\maketitle

\begin{center}
\footnotesize
\raisebox{-0.2em}{\includegraphics[height=1.15em]{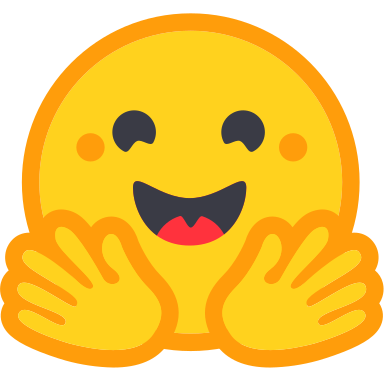}}\,\textbf{Hugging Face}\quad
\href{https://huggingface.co/cccsds/openha-qwen3vl8b-coldstart-step16200}{\textcolor{blue}{SFT weights}}\quad$\vert$\quad
\href{https://huggingface.co/cccsds/openha-qwen3vl8b-grpo-step400}{\textcolor{blue}{GRPO weights}}\quad$\vert$\quad
\href{https://huggingface.co/datasets/cccsds/openha-minecraft-eval-149}{\textcolor{blue}{Test set (149 tasks)}}
\end{center}

\begin{abstract}
A failed gaming trajectory reveals that an agent needs intervention,
but not necessarily how its execution harness should change.
Effective harness adaptation therefore requires gathering diagnostic
evidence, rather than merely proposing alternative recovery rules.
We present KPHarness, a framework that adapts the external execution
harness of a frozen vision-language-action model through
knowledge--probing co-evolution.
Recovery knowledge supplies context-specific intervention hypotheses
that guide active probing; probe outcomes reveal when these hypotheses
apply and provide evidence for refining both recovery knowledge and
subsequent probe selection.
This reciprocal process allows the harness to improve not only which
interventions it applies, but also how it investigates failures.
KPHarness separates fast policy execution from a slower research loop
that translates accumulated knowledge and probing evidence into bounded
harness revisions.
Each revision is compared with the incumbent under matched sandbox
conditions and accepted only when the paired evaluation shows improvement.
Recoveries supported by further validation are retained for cross-task
reuse, while immediate interaction feedback and delayed episode-success
credit update action- and knowledge-probing policies during adaptation.
Experiments on Minecraft tasks spanning mining, combat, and crafting
show improved task coverage over bare policy execution, a manually
specified fixed harness, and knowledge-only harness evolution with
non-adaptive probing.
Evaluation on previously unseen tasks further shows improved task
completion with the evolved harness, knowledge bank, and probing policies
held fixed, without additional search or persistent updates.
These findings support knowledge--probing co-evolution as a mechanism
for turning interaction experience into transferable harness improvements
without further updating the underlying actor. The project page is available at
\url{https://chenzk202212.github.io/adaptive-gaming-harnesses/}.
\end{abstract}

\section{Introduction}
\label{sec:introduction}

Vision-language-action (VLA) models
\citep{brohan2023rt, kim2024openvla, chen2025combatvla, li2025jarvis,
wang2026unified} connect visual observations and language instructions with
low-level control, enabling increasingly capable agents in interactive
environments. Long-horizon gaming, however, exposes limitations beyond
single-step action prediction. Even a capable VLA may repeatedly execute an
ineffective action, fail to recover from a stalled trajectory, or overlook
experience relevant to the current failure. As interaction horizons grow,
performance therefore depends not only on the capabilities encoded in model
parameters, but also on the external harness that governs how the model acts,
probes, remembers, and recovers.

Existing work improves long-horizon VLA execution through prompting, memory,
recovery, and hierarchical control
\citep{li2025towards, fan2025long, yang2025lohovla}, but these mechanisms are
often manually designed and fixed at deployment. Recent automatic harness
methods instead search for prompts, tools, memory, and control logic from
execution feedback
\citep{lee2026meta, chen2026harnessx, lou2026autoharness}. Such methods typically evaluate harness modifications through their execution outcomes. Visual gaming introduces an additional challenge: a failed trajectory often reveals what happened but not why. For example, an unsuccessful attack may result from target loss, insufficient range, an obstacle, or insufficient persistence, each requiring different recovery knowledge. Consequently, repeatedly trying harness modifications is inefficient when the underlying failure remains ambiguous. \method{} turns automatic harness evolution from outcome-driven search into evidence-guided evolution through knowledge–probing co-evolution: evolving recovery knowledge provides context-specific intervention hypotheses that can focus subsequent probing, while probe outcomes determine their applicability and provide evidence for refining both knowledge and future probe selection. These challenges motivate our central question:

\begin{figure*}[t]
\centering
\includegraphics[width=0.92\textwidth]{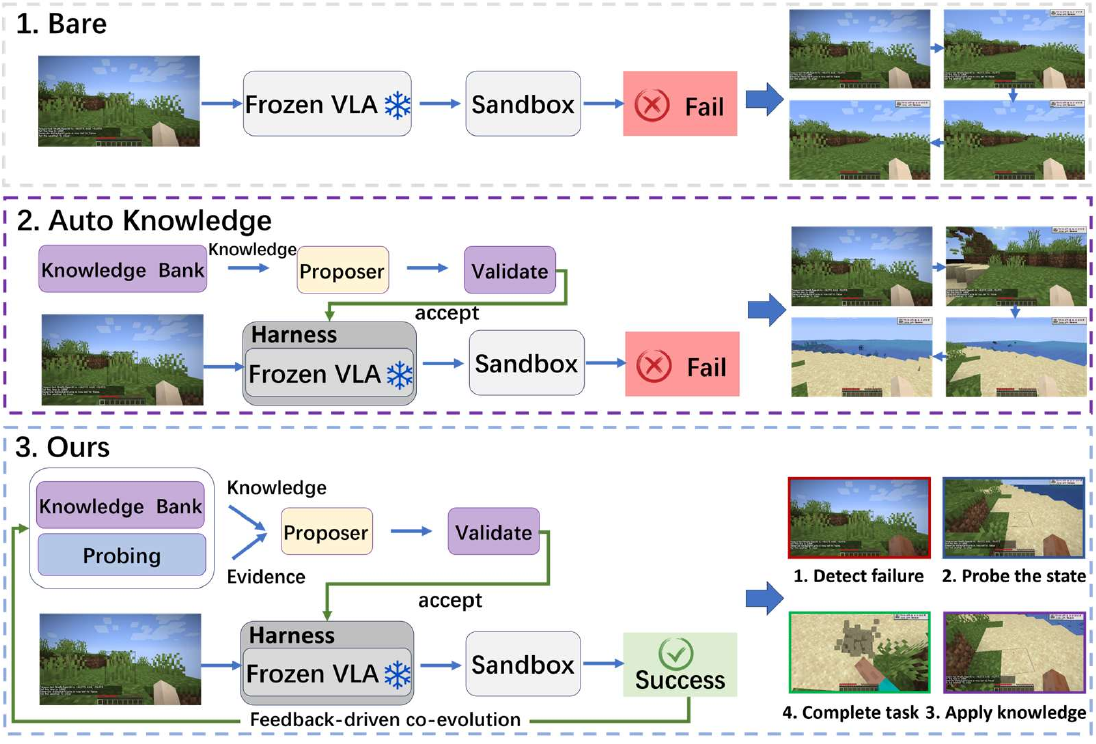}
\vspace{-.3cm}
\caption{
\emph{Knowledge--probing co-evolution} around a shared frozen VLA.
Auto-K proposes and validates harness modifications using accumulated
knowledge, while \method{} additionally uses probing evidence and interaction
feedback to adapt probe and knowledge selection. In the illustrated
sand-mining example, Bare and Auto-K fail, whereas \method{} recovers and
completes the task.
}
\label{fig:framework}
\vspace{-.6cm}
\end{figure*}

\begin{tcolorbox}[
    colback=white,
    colframe=gray,
    arc=3mm,
    boxrule=0.8pt,
    shadow={0.5mm}{-0.5mm}{0mm}{black!50!white}
]
\itshape
Can transferable recovery knowledge guide what a harness should probe under
ambiguous failures, while probe outcomes identify applicable knowledge and
refine subsequent probing and knowledge selection?
\end{tcolorbox}


Existing harness search proposes modifications primarily from execution outcomes. In visually ambiguous environments, however, the same failure outcome may admit multiple incompatible explanations. We therefore treat harness evolution as a sequential process of hypothesis formation, active diagnosis, and empirical verification, where transferable knowledge supplies recovery hypotheses, and probing determines which hypothesis applies. 
We address this question with \method{}, a framework for
\textit{test-time harness evolution} around a frozen VLA. Here,
\emph{test-time} means that the VLA parameters remain fixed after training,
while only the external harness evolves through environment interaction. At
the fast timescale, the VLA executes actions under the current harness. When
execution stalls, the harness selects a probe according to the task and
failure context, executes it in the environment, and uses the resulting
evidence to retrieve relevant recovery knowledge. At the slow timescale, a
vision-language researcher converts this evidence into task-specific harness
modifications. Candidates are compared with their incumbents through paired
sandbox evaluation, and accepted modifications are subsequently assessed on
disjoint holdout seeds. Immediate interaction utility and delayed
episode-success credit are then used to update probe and knowledge selection.
Thus, unlike outcome-driven harness search alone, \method{} learns both
\textit{which intervention to apply} and \textit{what to probe in order to
discover it}.

This process creates a knowledge--probing co-evolution loop. Better probes
provide clearer evidence for selecting and refining interventions, while
successful interventions reveal which probes and knowledge are useful under
particular task and failure contexts. Validated recoveries are distilled into
transferable execution knowledge and reused on subsequent tasks, while
unresolved failures and probing ambiguities are converted into targeted
follow-up cases for further sandbox evaluation. As illustrated in
Figure~\ref{fig:framework}, \emph{Auto-K} searches for harness modifications
from accumulated knowledge but does not improve its probing policy from probe
outcomes. In contrast, \method{} uses subsequent interaction to update both
probe and knowledge selection. Experiments on 149 Minecraft tasks show that
harness evolution improves coverage over bare execution and a fixed harness,
while the complete framework further improves over knowledge-only evolution
without updating the underlying VLA.

Our contributions are fourfold.
\begin{enumerate}[
leftmargin=*,
topsep=2pt,
itemsep=1pt,
parsep=0pt,
partopsep=0pt
]

\item \emph{Evidence-guided test-time harness evolution.}
We identify a limitation of outcome-driven automatic harness search in visually ambiguous environments and formulate harness evolution as an evidence-guided process around a frozen VLA.

\item \emph{Knowledge--probing co-evolution for evidence-guided search.}
We couple transferable recovery knowledge with active probing: knowledge provides intervention hypotheses, probes gather evidence about their applicability, and validated execution outcomes update both future knowledge and probe selection.

\item \emph{Sandbox-grounded accumulation and transfer.}
Candidate harness changes are accepted only through paired environment interaction, while validated recoveries are promoted into transferable knowledge that guides subsequent tasks.

\item \emph{Probe-driven failure-frontier expansion.}
We convert failure reflections and probing ambiguities into targeted follow-up
cases, exposing new failure modes that guide subsequent harness evolution. 

\end{enumerate}

\vspace{-0.4cm}
\section{Related Work}
\vspace{-0.25cm}

\textbf{Static harnesses for gaming and embodied agents.}
Prior work has advanced Minecraft agents through video pretraining,
language-conditioned behavior learning, and unified development frameworks
\citep{fan2022minedojo,baker2022video,lifshitz2023steve,
cai2024minestudio,li2025jarvis}. However, gaming VLAs typically produce short
action segments and may lose targets, repeat ineffective actions, or fail to
verify progress. External harnesses mitigate these limitations through task
state, memory, action constraints, verification, and recovery. Recent embodied harnesses extend frozen VLA and world-action models through
cross-stage context \citep{zeng2026helm}, reusable primitives
\citep{zhang2026harness}, and fast--slow control
\citep{gu2026harnesswam}. Yet their intervention and recovery mechanisms are
largely fixed before deployment. Our method instead probes ambiguous failures,
acquires transferable recovery knowledge, and improves subsequent probing and
knowledge selection while keeping the gaming VLA frozen.

\textbf{Automatic harness evolution.}
Early self-evolving agents primarily optimized prompts. APE and OPRO treat
instruction generation as black-box optimization
\citep{zhou2022large,yang2024large}; EvoPrompt and Promptbreeder use
evolutionary search \citep{guo2024connecting,fernando2023promptbreeder};
and ProTeGi and TextGrad guide revisions with textual feedback
\citep{pryzant2023automatic,yuksekgonul2024textgrad}. DSPy and MIPRO extend
optimization to instructions and demonstrations in multi-stage LM programs
\citep{khattab2023dspy,opsahl2024optimizing}. Recent methods further optimize
broader harness components, including tools, memory, middleware, and control
logic \citep{lee2026meta,chen2026harnessx,lin2026agentic,lou2026autoharness}. These methods mainly target coding, textual, or programmatically verifiable
environments with explicit execution feedback. Long-horizon gaming instead
involves visually and temporally ambiguous failures, and proposed
modifications may be difficult for a frozen VLA to execute. \method{} therefore
probes the environment to identify failure contexts and applicable recovery
knowledge. Interaction feedback improves subsequent probe and knowledge
selection, while validated recoveries transfer across tasks. This
knowledge--probing co-evolution improves both \emph{what intervention to apply}
and \emph{what evidence to seek}.


\section{Preliminaries}
\label{sec:preliminaries}

\subsection{Long-Horizon Gaming with a Frozen VLA}
\label{sec:prelim_vla}

We consider a visual gaming environment in which an agent completes a task
specified by a language instruction \(g\). At environment timestep \(t\), the
agent receives a visual observation \(o_t\), executes a low-level action
\(a_t\), and observes the next observation \(o_{t+1}\), an optional reward
\(r_t\), and task-progress or failure signals. The available interaction
history is
\[
h_t = (o_{\leq t},a_{<t},r_{<t}),
\]
and an episode induces a trajectory
\[
\tau =
(o_1,a_1,r_1,\ldots,o_T,a_T,r_T,o_{T+1}),
\]
whose task-level outcome is measured by an environment-defined objective
\(R(\tau,g)\).

Low-level action proposals are generated by a pretrained
vision-language-action policy \(\pi_\theta\). We study the
\emph{frozen-actor} setting, in which the parameters \(\theta\) remain
unchanged throughout execution, harness adaptation, and cross-task transfer.
Any improvement must therefore arise from the external execution process.
For long-horizon gaming, this process must preserve relevant context,
recognize stalled progress, diagnose visually ambiguous failures, and recover
through subsequent interaction.

\subsection{Test-Time Harness Evolution}
\label{sec:prelim_harness}
\label{sec:prelim_evolution}

A \emph{test-time harness} \(\mathcal{H}_\phi\) is a model-external
execution scaffold with configuration \(\phi\) and external state
\(\mathcal{M}\). Given the task and interaction history, the harness
constructs an execution context, queries the frozen actor, and determines
the action executed in the environment:
\begin{align}
c_t
&=
\mathcal{C}_\phi(h_t,g,\mathcal{M}),
\label{eq:harness_context}\\
\tilde{a}_t
&\sim
\pi_\theta(\cdot\mid c_t),
\label{eq:actor_proposal}\\
a_t
&=
\mathcal{I}_\phi
(h_t,g,\mathcal{M},\tilde{a}_t).
\label{eq:harness_intervention}
\end{align}
Here, \(\mathcal{C}_\phi\) assembles the context supplied to the actor,
while \(\mathcal{I}_\phi\) may pass through, replace, or block the actor's
proposal, including interventions for probing and recovery. The configuration
\(\phi\) specifies mechanisms such as intervention triggers, recovery
instructions, verification rules, and memory-use policies. The external state
\(\mathcal{M}\) stores accumulated experience, reusable recovery knowledge,
and statistics used for probe and knowledge selection.

A static harness may retrieve different memories or produce different actions,
but it does so using mechanisms fixed in advance. Harness evolution
additionally allows these mechanisms and their adaptive external state to
change from interaction evidence. Let
\(P_{\theta,\phi,\mathcal{M}}(\tau\mid g)\) denote the trajectory distribution
induced jointly by the environment, frozen actor, and harness. At adaptation
stage \(k\), the current system collects task-conditioned interaction evidence
\begin{equation}
\begin{aligned}
\mathcal{D}_k
&=
\left\{(g_i,\tau_i)\right\}_{i=1}^{N_k},\\
\tau_i
&\sim
P_{\theta,\phi_k,\mathcal{M}_k}(\cdot\mid g_i).
\end{aligned}
\label{eq:evolution_data}
\end{equation}
An abstract update operator \(\mathcal{U}\) then revises the harness while
leaving the actor unchanged:
\begin{equation}
\begin{aligned}
(\phi_{k+1},\mathcal{M}_{k+1})
&=
\mathcal{U}
(\phi_k,\mathcal{M}_k,\mathcal{D}_k),\\
\theta_{k+1}
&=
\theta_k
=
\theta.
\end{aligned}
\label{eq:harness_update}
\end{equation}
The index \(k\) denotes harness-adaptation stages and is distinct from the
environment timestep \(t\). An adaptation stage may include multiple sandbox
episodes, and the formulation does not prescribe a particular update schedule
or learning rule.

Given a deployment-task distribution \(p_{\mathrm{dep}}\), the objective is
to improve the expected performance of the complete actor--harness system:
\begin{equation}
\max_{\phi_K,\mathcal{M}_K}
\;
\mathbb{E}_{\substack{
g\sim p_{\mathrm{dep}},\\
\tau\sim
P_{\theta,\phi_K,\mathcal{M}_K}(\cdot\mid g)
}}
\left[
R(\tau,g)
\right].
\label{eq:harness_objective}
\end{equation}
We refer to this setting as \emph{test-time harness evolution}: after actor
training, the harness adapts through repeated environment interaction,
potentially across multiple episodes and tasks, while the underlying VLA
parameters remain frozen.

\section{Methodology}
\label{sec:method}
\begin{figure*}[t]
\centering
\includegraphics[width=\textwidth]{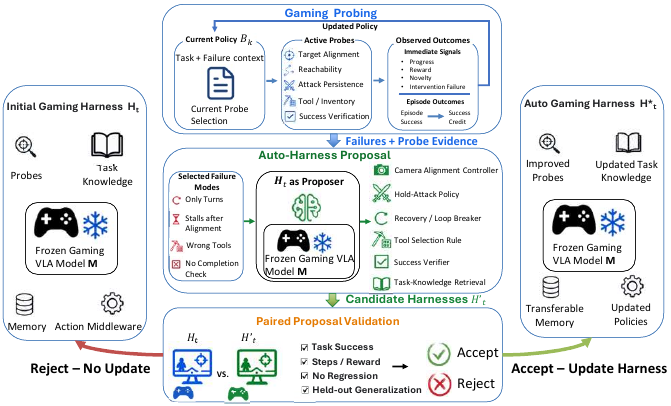}
\vspace{-.3cm}
\caption{
\emph{Automatic harness evolution through knowledge--probing co-evolution.}
Action probes collect evidence about ambiguous failures, while knowledge
probes test which recovery strategies apply. Probe evidence and accumulated
knowledge guide harness proposals, which are compared with the incumbent
through paired sandbox validation. Interaction outcomes improve both probing
policies, and validated recoveries become transferable knowledge, while the VLA remains frozen.
}
\label{fig:method}
\vspace{-.4cm}
\end{figure*}

\method{} (see Figure~\ref{fig:method} ) operationalizes evidence-guided harness evolution through knowledge–probing co-evolution. Rather than proposing harness modifications from terminal outcomes alone, the framework first probes ambiguous failures, uses the resulting evidence to test applicable recovery knowledge, and then proposes evidence-grounded modifications.  The frozen VLA
acts in a fast execution loop, while a vision-language researcher performs
diagnosis, proposal, and verification outside this latency-sensitive loop.
Subsequent interaction outcomes update both action and knowledge probing, and
validated recoveries become reusable knowledge for later tasks.

\subsection{Adaptive Gaming Harness}
\label{sec:adaptive_gaming_harness}

We follow the notation in Section~\ref{sec:preliminaries}. At evolution
round \(k\), the incumbent harness has configuration \(\phi_k\) and external
state \(\mathcal{M}_k\). The external state contains episodic experience, a
recovery-knowledge bank \(\mathcal{K}_k\), an action-probing policy
\(\mathcal{B}^{a}_k\), and a knowledge-probing policy
\(\mathcal{B}^{z}_k\). Given the current task and failure context, the harness
constructs or specializes candidate probes from the fixed low-level action
space, and the action-probing policy selects among them to reveal additional
environment evidence. The knowledge-probing policy selects a stored recovery
strategy as a hypothesis, specializes it to the current task, and tests its
applicability through subsequent interaction. Knowledge probing is therefore
not passive retrieval: its execution outcome determines whether the selected
knowledge should be preferred in similar failure contexts. The index \(k\)
identifies evolution rounds, although both probing policies may also improve
during learning rollouts within the same round.

At each environment step, the harness retrieves relevant experience,
constructs the actor's context, and regulates the proposed action as defined
in Equations~\ref{eq:harness_context}--\ref{eq:harness_intervention}.
Its editable components include intervention triggers, recovery instructions,
tool-selection rules, memory retrieval, completion checks, and probing
behavior. Probe outcomes are recorded during interaction, episode outcomes
provide delayed feedback, and the accumulated evidence supports slower harness
revision. The VLA parameters and low-level action space remain fixed
throughout this process.

\subsection{Adaptive Gaming Probing}
\label{sec:gaming_weakness_probing}

Probing is activated when progress stalls or execution repeatedly fails. It
targets execution-relevant uncertainties such as target alignment, interaction
reachability, attack persistence, tool selection, and completion checks. Given
the current task and failure context, the harness first constructs or
specializes applicable action probes from the fixed low-level action space.
The action-probing policy then selects among these candidates and executes the
selected probe to gather additional environment evidence. The resulting
observation refines failure attributes such as target loss, orientation error,
insufficient reachability, or repeated stalling. Conditioned on this refined
failure context, the knowledge-probing policy selects a recovery strategy from
\(\mathcal{K}_k\), adapts it to the current task, and tests its applicability
through subsequent interaction.

Both probing policies maintain separate statistics and use a
context-conditioned, UCB-inspired selection rule
\citep{auer2002finite,li2010contextual}. The broader design is motivated by
information-seeking exploration, cross-context transfer, and hierarchical
decision making
\citep{houthooft2016vime,liu2018transferable,hong2022hierarchical}.
For either policy, let \(x\) denote the task and failure context and \(v\) an
available action probe or recovery strategy. For action probing, the available
candidates are task-conditioned and may include newly instantiated or
specialized probes. Let \(N_{x,v}\) count previous selections of \(v\) in
context \(x\), \(Q_{x,v}\) denote their cumulative utility, and
\(\bar{q}_v\) denote the mean utility of \(v\) across contexts. The prior
weight \(\lambda>0\) controls cross-context transfer, while \(\beta\geq0\)
controls exploration. Each policy selects the candidate with the highest
score:
\vspace{-.1cm}
\begin{equation}
\operatorname{UCB}(v\mid x)
=
\frac{Q_{x,v}+\lambda\bar{q}_v}
     {N_{x,v}+\lambda}
+
\beta
\sqrt{
\frac{
\log\!\left(1+\sum_{v'}N_{x,v'}\right)
}{
N_{x,v}+1
}
}.
\label{eq:contextual_ucb}
\end{equation}
\vspace{-.1cm}

The first term combines context-specific experience with the global value of
the candidate, allowing evidence to transfer across related contexts. The
second term encourages exploration of less frequently selected candidates.
More detailed derivations are provided in
Appendix~\ref{app:tthe-theory}.

During learning rollouts, each selected probe receives immediate feedback from
the environment. For a probe selected at timestep \(t\), let
\(\Delta p_t\) denote progress change, \(r_t\) environment reward,
\(n_t\) interaction novelty, and \(\mathrm{fail}_t\) an intervention
failure. Let \(c_p\) be the probe cost and
\(w_p,w_r,w_n,w_f\) the corresponding nonnegative weights. The immediate
utility is
\begin{equation}
q_t^{\mathrm{imm}}
=
w_p\Delta p_t
+
w_r r_t
+
w_n n_t
-
c_p
-
w_f\mathbb{I}[\mathrm{fail}_t],
\label{eq:immediate_probe_utility}
\end{equation}
where \(\mathbb{I}[\cdot]\) is the indicator function. Action probes use the
novelty term to reward newly revealed interaction evidence. Knowledge probes
may set \(w_n=0\) and instead evaluate the progress produced after applying
the selected recovery strategy.

Because an informative probe may affect the final outcome only after several
subsequent actions, we additionally assign heuristic delayed credit after
successful episodes. Let \(\mathrm{Succ}(\tau,g)\) indicate successful
completion of task \(g\), \(R_{\mathrm{succ}}\) be the success-bonus
magnitude, \(d_t\) the temporal distance from the probe to episode completion,
and \(\gamma\in(0,1]\) a discount factor. The delayed bonus is
\begin{equation}
q_t^{\mathrm{delayed}}
=
R_{\mathrm{succ}}\gamma^{d_t}
\mathbb{I}[\mathrm{Succ}(\tau,g)].
\label{eq:delayed_probe_credit}
\end{equation}

Unsuccessful episodes receive no delayed success bonus. Each probe selection
increments its count once, while its cumulative utility incorporates both
immediate feedback and delayed success credit. These updates adapt
\(\mathcal{B}^{a}_k\) and \(\mathcal{B}^{z}_k\) from interaction experience
without updating the VLA.

The harness also records the observations, interventions, progress changes,
and outcomes associated with each probe. Related records are grouped into
recurring failure modes, such as turning without reacquiring the target,
stalling after alignment, acting outside interaction range, using an
incorrect tool, or missing a completion check. These records provide evidence
for subsequent harness proposals.

\subsection{Evidence-Grounded Harness Proposal}
\label{sec:auto_harness_proposal}

A vision-language researcher serves as the proposer. It receives the
incumbent harness, selected failure modes and their supporting probe records,
and relevant recovery knowledge from \(\mathcal{K}_k\). Rather than inferring
a modification from the failed trajectory alone, the researcher uses probe
evidence to connect a likely failure cause with an appropriate intervention.
It returns a failure diagnosis, an evidence-linked rationale, and a bounded
edit to the structured \textsc{harness configuration}. Applying the edit to
\(\phi_k\) produces a candidate configuration \(\tilde{\phi}_k\).

Candidate edits may revise camera alignment, attack persistence,
reachability handling, loop-breaking behavior, tool selection, completion
checks, or knowledge retrieval. The researcher can modify only components
exposed by the \textsc{harness configuration}; it cannot update the VLA,
replace the actor, modify the environment, or alter the task objective. A
candidate therefore remains a hypothesis until sandbox interaction
demonstrates that the frozen actor can execute it effectively.

\subsection{Paired Validation and Knowledge Transfer}
\label{sec:paired_validation}

The incumbent and candidate are compared on the same task--seed pairs
\(\mathcal{V}_k\), where \(g\) denotes a task and \(s\) an environment
seed. Both conditions start from the same external-state snapshot, including
episodic memory and probing-policy statistics. Persistent memory and policy
writeback are disabled during paired and holdout evaluation.

For a pair \((g,s)\), let \(J_{\mathrm{cand}}(g,s)\) and
\(J_{\mathrm{inc}}(g,s)\) denote the candidate and incumbent scores under a
fixed evaluation rule. The score prioritizes task success, with progress,
reward, execution cost, and regression as secondary criteria. The average
paired improvement is
\begin{equation}
\widehat{\Delta J}_k
=
\frac{1}{|\mathcal{V}_k|}
\sum_{(g,s)\in\mathcal{V}_k}
\left[
J_{\mathrm{cand}}(g,s)
-
J_{\mathrm{inc}}(g,s)
\right].
\label{eq:paired_validation}
\end{equation}
Given an acceptance threshold \(\epsilon_{\mathrm{pair}}\), the next harness
configuration is
\begin{equation}
\phi_{k+1}
=
\begin{cases}
\tilde{\phi}_k,
& \widehat{\Delta J}_k \geq \epsilon_{\mathrm{pair}},\\
\phi_k,
& \text{otherwise}.
\end{cases}
\label{eq:accept_reject}
\end{equation}
Rejected candidates leave the incumbent unchanged. Accepted candidates are
subsequently evaluated on disjoint holdout seeds. These results do not alter
the original acceptance decision, but provide evidence of cross-seed transfer.

An accepted recovery is stored as a candidate knowledge record containing its
triggering context, recovery strategy, and supporting evidence, with
task-specific details removed. It is promoted to transferable knowledge when
it improves performance on a holdout seed or succeeds again on another task.
Promoted records enter \(\mathcal{K}_{k+1}\) and are tested by the
knowledge-probing policy in subsequent contexts. Outside evaluation, learning rollouts update probing statistics and episodic
experience. The selected harness, probing policies, and promoted knowledge
then initialize subsequent tasks, forming the knowledge--probing co-evolution
loop. Uncertain diagnoses and inconsistent recovery outcomes are additionally
converted into validated, deduplicated follow-up cases, which remain separate
from the fixed evaluation set and target the evolving failure frontier.

\section{Experiments}
\label{sec:experiments}
\vspace{-.2cm}

\begin{figure*}[h!]
\centering
\includegraphics[width=\textwidth]{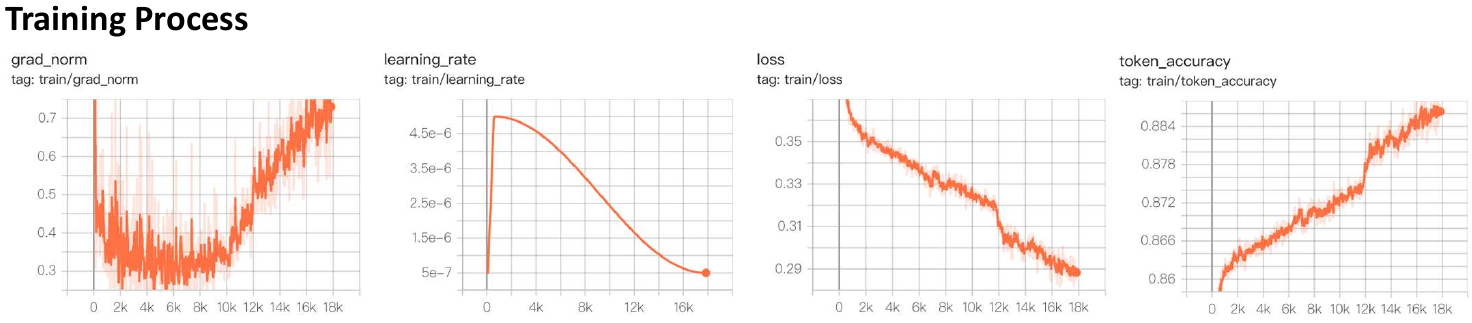}
\vspace{-.5cm}
\caption{
Training curves for cold-start supervised fine-tuning (SFT), showing
decreasing loss and improving token accuracy.
}
\vspace{-.1cm}
\label{fig:sft}
\end{figure*}

\begin{figure*}[h!]
\centering
\includegraphics[width=\textwidth]{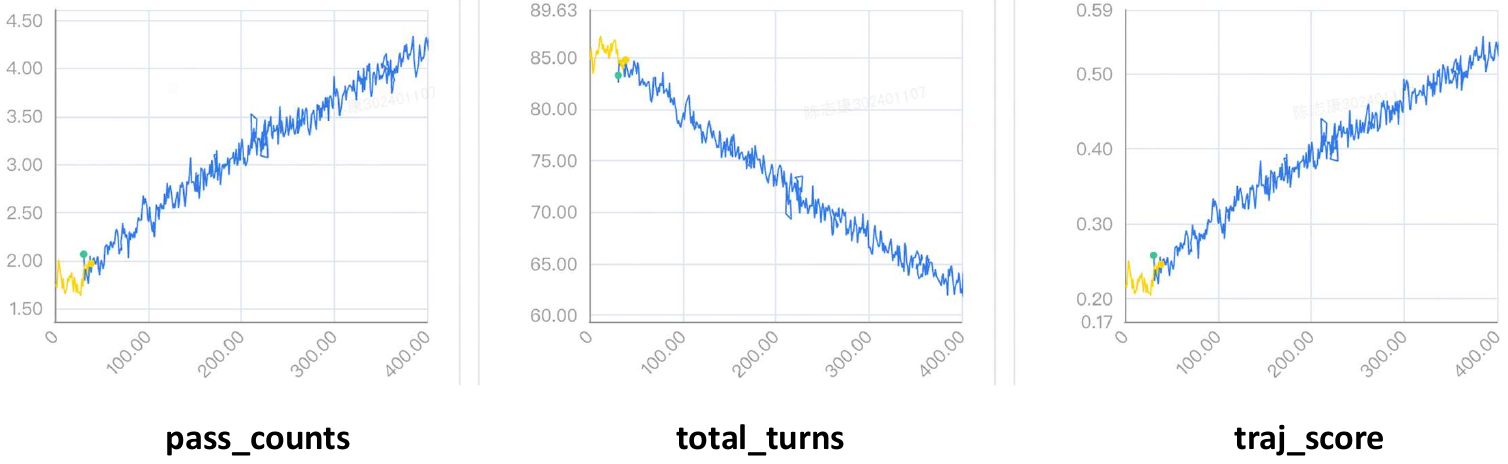}
\vspace{-.5cm}
\caption{
GRPO training dynamics over 400 optimization steps on 32 interactive OpenHA
training tasks in Minecraft, initialized from the SFT checkpoint at step
16,200. Pass counts and trajectory scores increase, while total interaction
turns decrease.
}
\label{fig:grpo}
\end{figure*}

We first train a gaming VLA based on Qwen3-VL-8B-Instruct
\citep{bai2025qwen3} through cold-start supervised fine-tuning (SFT), with
the training configuration detailed in Appendix~\ref{setup}. Starting from
the SFT checkpoint at step 16,200, we perform 400 steps of Group Relative
Policy Optimization (GRPO) \citep{shao2024deepseekmath} on 32 interactive
OpenHA training tasks in Minecraft. Figures~\ref{fig:sft}
and~\ref{fig:grpo} show the corresponding training curves. We then freeze
the resulting policy and evaluate harness adaptation on 149 OpenHA tasks
spanning mining, combat, and crafting.

The slow research loop uses Gemini-2.5-Flash
\citep{comanici2025gemini} at temperature \(0\) as the vision-language
researcher and harness proposer. It receives the incumbent harness,
structured failure and probe records, relevant recovery knowledge, and up
to four frames sampled from the failed trajectory, and returns a structured
candidate patch. Each task uses three proposal rounds with at most one
candidate per round, and Auto-K and \method{} use the same proposer
configuration. Candidate patches are not accepted according to the
researcher's judgment alone; they are executed and verified in our locally
deployed MineExplorer environment \citep{ju2026mineexplorer}. Accepted
candidates are subsequently evaluated on a disjoint holdout seed. The
researcher neither generates low-level actions nor participates in the
latency-sensitive execution loop.

\subsection{Evaluation Setup}
\label{sec:evaluation_setup}

We evaluate on 149 OpenHA tasks in MineExplorer, including 50 mining,
49 combat, and 50 crafting tasks, using the same frozen GRPO checkpoint
for all methods. Each episode allows up to 100 VLA decisions, with four
\(50\,\mathrm{ms}\) low-level actions per decision, yielding up to 400
temporally dependent actions per rollout.

Our primary setting uses the \texttt{legacy} protocol with sampling
temperature \(0.7\), where each prediction conditions only on the current
observation and task instruction. We additionally evaluate Bare and Fixed
under \texttt{mtv1} at temperature \(0\), which restores the latest ten
observation--action turns. Both protocols predict a \(200\,\mathrm{ms}\)
action chunk per decision. Action definitions and prompt construction are
provided in Appendix~\ref{experiments}.

We use one probing configuration across all tasks and seeds:
\(w_p=1.0\), \(w_r=0.25\), \(w_n=0.15\), \(c_p=0.05\),
\(w_f=0.5\), \(\beta=1.2\), \(\lambda=0.25\),
\(R_{\mathrm{succ}}=1.0\), and \(\gamma=0.8\).
Knowledge probing uses the same coefficients, with cost \(0.02\) and no
novelty term. Under the primary setting, Bare solves 39 of 149 tasks; the remaining 110
form the \emph{Bare-failure adaptation subset}. Candidate and incumbent
harnesses are compared using the same task, search seed, decoding settings,
memory snapshot, and probing-policy state. A candidate is accepted when its
paired score improves by at least \(0.01\); the exact scoring rule is given in
Appendix~\ref{sec:harness_search_details}. After adaptation, we freeze the harness and probing statistics and replay all
149 tasks without further search or writeback. Frozen probes may still execute,
but their policies are not updated. Because this replay includes tasks seen
during adaptation, we report it as \emph{task coverage}, while disjoint-seed
holdout evaluation measures cross-seed robustness.

\vspace{-.4cm}

\vspace{-.5cm}

\subsection{Compared Methods and Main Results}
\label{sec:main_results}

We compare four settings using the same frozen VLA checkpoint.
\emph{Bare} directly executes the VLA, while \emph{Fixed} adds manually
specified retry and recovery rules. \emph{Auto Knowledge (Auto-K)} further
searches for and validates recovery knowledge through sandbox interaction,
but uses predefined or coverage-based probing. The full \method{} additionally
adapts probe construction and action- and knowledge-probe selection from
interaction feedback. None of the methods updates the VLA parameters. Recent automatic harness methods such as Meta-Harness~\citep{lee2026meta}
primarily study textual, verifiable environments, whereas visual gaming
introduces ambiguous failures that often require interaction to diagnose.
Our variants therefore progress from fixed recovery, to automatic knowledge
accumulation, and finally to knowledge--probing co-evolution.

For failed trajectories, Auto-K and \method{} use the same proposer to
generate bounded harness patches, which are accepted only when they improve
over the incumbent under paired evaluation and are subsequently tested on a
disjoint holdout seed. An \emph{HO-win} denotes improvement on this holdout
evaluation. The key difference is adaptive probing: \method{} updates
contextual action- and knowledge-probing policies using immediate interaction
feedback and delayed episode-success credit, whereas Auto-K does not learn
from probe outcomes.


\begin{table*}[h!]
\centering
\vspace{-.2cm}
\caption{
Overall task coverage on the 149-task OpenHA evaluation set in Minecraft.
}
\vspace{-.2cm}
\label{tab:openha_results}
\resizebox{\textwidth}{!}{
\begin{tabular}{l|c|c|c|ccc|c|cc|cc}
\hline
Method
& Protocol
& Temp.
& Overall
& \multicolumn{3}{c|}{Task success}
& Rescued
& \multicolumn{2}{c|}{Knowledge search}
& \multicolumn{2}{c}{Adaptive probing} \\
\cline{5-7}\cline{9-10}\cline{11-12}
&
&
&
& Mine & Kill & Craft
&
& ACCEPT & HO-win
& P-only & Replay succ. \\
\hline

Bare
& mtv1
& \(0\)
& 26/149 (17.4\%)
& 9/50 & 17/49 & 0/50
& 0
& -- & --
& -- & -- \\

Fixed
& mtv1
& \(0\)
& \textbf{34/149 (22.8\%)}
& \textbf{16/50} & \textbf{17/49} & \textbf{1/50}
& \textbf{8}
& -- & --
& -- & -- \\

\hline

Bare
& legacy
& \(0\)
& 23/149 (15.4\%)
& 8/50 & 15/49 & 0/50
& 0
& -- & --
& -- & -- \\

Fixed
& legacy
& \(0\)
& 31/149 (20.8\%)
& 13/50 & 18/49 & 0/50
& 8
& -- & --
& -- & -- \\

Auto-K
& legacy
& \(0\)
& 41/149 (27.5\%)
& 20/50 & 20/49 & 1/50
& 18
& 11 & 3
& N/A & N/A \\

\rowcolor{blue!8}
\method{}
& legacy
& \(0\)
& \textbf{44/149 (29.5\%)}
& \textbf{21/50} & \textbf{21/49} & \textbf{2/50}
& \textbf{21}
& 14 & 3
& 1 & 1 \\

\hline

Bare
& legacy
& \(0.7\)
& 39/149 (26.2\%)
& 20/50 & 18/49 & 1/50
& 0
& -- & --
& -- & -- \\

Fixed
& legacy
& \(0.7\)
& 50/149 (33.6\%)
& 23/50 & 25/49 & 2/50
& 11
& -- & --
& -- & -- \\

Auto-K
& legacy
& \(0.7\)
& 63/149 (42.3\%)
& 25/50 & 30/49 & 8/50
& 24
& 14 & 5
& N/A & N/A \\

\rowcolor{blue!8}
\method{}
& legacy
& \(0.7\)
& \textbf{70/149 (47.0\%)}
& \textbf{29/50} & \textbf{33/49} & \textbf{8/50}
& \textbf{31}
& 14 & 5
& \textbf{3} & \textbf{17} \\

\hline
\end{tabular}

}

\vspace{1mm}
\begin{minipage}{0.99\textwidth}
\footnotesize
Rescued counts tasks failed by Bare but solved by the corresponding
method under the same protocol and temperature.
``N/A'' indicates that the method does not contain the adaptive
probing and delayed-credit mechanism required to define probe-only
or probing-replay success.
\end{minipage}
\end{table*}

\begin{table*}[h!]
\centering
\vspace{-.3cm}
\caption{
Multi-seed evaluation on the same 110 tasks failed by Bare,
using the \texttt{legacy} protocol with temperature \(0.7\).
}
\label{tab:multiseed_robustness}
\resizebox{\textwidth}{!}{
\begin{tabular}{l|ccc|c|c|cc|cc}
\hline
Method
& \multicolumn{3}{c|}{Task success}
& Rescued
& Success rate
& \multicolumn{2}{c|}{Knowledge search}
& \multicolumn{2}{c}{Adaptive probing} \\
\cline{2-4}\cline{7-8}\cline{9-10}
& Mine & Kill & Craft
&
&
& ACCEPT & HO-win
& P-only & Replay succ. \\
\hline

Auto-K
& \(4.7_{\scriptscriptstyle\pm 0.6}/30\)
& \(7.7_{\scriptscriptstyle\pm 5.1}/31\)
& \(3.3_{\scriptscriptstyle\pm 3.2}/49\)
& \(15.7_{\scriptscriptstyle\pm 8.0}/110\)
& \(14.2_{\scriptscriptstyle\pm 7.3}\%\)
& -- & --
& N/A & N/A \\

\rowcolor{blue!8}
\method{}
& \(\mathbf{8.3}_{\scriptscriptstyle\pm 1.2}/30\)
& \(\mathbf{13.7}_{\scriptscriptstyle\pm 1.5}/31\)
& \(\mathbf{6.7}_{\scriptscriptstyle\pm 0.6}/49\)
& \(\mathbf{28.7}_{\scriptscriptstyle\pm 3.2}/110\)
& \(\mathbf{26.1}_{\scriptscriptstyle\pm 2.9}\%\)
& \(14.0_{\scriptscriptstyle\pm 3.0}\)
& \(6.7_{\scriptscriptstyle\pm 2.1}\)
& \(\mathbf{2.3}_{\scriptscriptstyle\pm 1.2}\)
& \(\mathbf{14.7}_{\scriptscriptstyle\pm 3.2}\) \\

\hline
\end{tabular}
}
\end{table*}





\vspace{-.1cm}
\textbf{Main Results and Cross-Seed Robustness.}
As shown in Table~\ref{tab:openha_results}, under the primary
\texttt{legacy} protocol with temperature \(0.7\), Bare, Fixed, and Auto-K
solve 39 (26.2\%), 50 (33.6\%), and 63 (42.3\%) of the 149 tasks,
respectively, while the full \method{} reaches 70 (47.0\%).
The progressive gains from Bare to Fixed, Auto-K, and \method{} support the
benefit of external recovery, transferable knowledge, and finally adaptive
knowledge--probing co-evolution. In particular, \method{} achieves 17
probing-enabled replay successes, including three strict probe-only
recoveries where Knowledge Search fails. The advantage is also consistent
across seeds: on the same 110 Bare-failed tasks, \method{} improves the mean
rescue count from
\(15.7_{\scriptscriptstyle\pm8.0}\) for Auto-K to
\(28.7_{\scriptscriptstyle\pm3.2}\), with positive gains under all three
evaluated seeds. Detailed per-seed results are provided in
Appendix~\ref{app:multiseed_robustness}.

\begin{wraptable}{l}{0.49\textwidth}
\vspace{-0.4cm}
\centering
\captionsetup{
  width=0.96\linewidth,
  justification=raggedright,
  singlelinecheck=false,
  font=small
}
\caption{Frozen transfer to 60 unseen tasks. All components are fixed during evaluation.}
\label{tab:frozen_transfer}

\vspace{-0.1cm}
\resizebox{0.96\linewidth}{!}{
\begin{tabular}{l|c|ccc}
\hline
Method & Overall & Mine & Kill & Craft \\
\hline
Bare
& 8/60 & 7/20 & 1/20 & 0/20 \\

Auto-K
& 7/60 & 6/20 & 1/20 & 0/20 \\

\rowcolor{blue!8}
\method{}
& \textbf{15/60}
& \textbf{9/20}
& \textbf{5/20}
& \textbf{1/20} \\
\hline
\end{tabular}
}
\vspace{-0.4cm}
\end{wraptable}

\textbf{Generalization to Unseen Tasks.}
We freeze the evolved harness, knowledge bank, and probing policy and evaluate
on 60 unseen tasks without further search or writeback. As shown in
Table~\ref{tab:frozen_transfer}, \method{} solves 15/60 tasks, compared with
8/60 for Bare and 7/60 for Auto-K, demonstrating transfer beyond the
adaptation set. Additional qualitative examples and learned recovery knowledge
are shown in Appendix Figures~\ref{fig:knowledge} and~\ref{fig:examples}.

\textbf{Ablation Study.}
Table~\ref{tab:openha_results} further separates the contributions of the
external harness components. Fixed consistently improves over Bare, while
Auto-K further improves over Fixed through transferable recovery knowledge.
The complete \method{} improves from 63 to 70 tasks over Auto-K, supporting
the additional benefit of adaptive probing and knowledge selection.
Complementary diagnostics in
Appendix~\ref{sec:static_knowledge_diagnostic} show that simply injecting a
fixed recovery bank does not reproduce the gain of the complete
sandbox-mediated evolution process. Additional analyses of knowledge search,
probe-only recovery, and sequential knowledge reuse are provided in
Appendix~\ref{sec:knowledge_probe_diagnostics} and~\ref{sec:sequential-knowledge-reuse}.

\vspace{-.3cm}
\section{Conclusion}
\label{sec:conclusion}
\vspace{-.1cm}

We introduced \method{}, a test-time harness-evolution framework that combines
adaptive probing, sandbox verification, and transferable recovery knowledge
around a frozen VLA. Experiments on Minecraft show improved task coverage and
transfer to unseen tasks without further actor training.

Several limitations remain. Due to the substantial computational cost of VLA
training and harness evolution, including SFT/GRPO training, repeated
environment rollouts, VLA inference, and multimodal researcher calls, our
evaluation is limited to a single policy backbone and gaming environment.
Harness evolution is also more expensive than fixed execution, while delayed
success credit remains heuristic and may not precisely identify the
contribution of individual probes. More detailed information can be found in Appendix \ref{sec:computational_cost}. Future work should improve adaptation
efficiency and evaluate knowledge--probing co-evolution across broader models
and task distributions.

\newpage
\appendix

\section{Experimental Setup}
\label{experiments}

\subsection{VLA Policy Training}
\label{setup}

We adopt Qwen3-VL-8B-Instruct as the policy backbone and train it in two
stages. First, we perform cold-start supervised fine-tuning (SFT) for three
epochs on approximately 765K multimodal Minecraft action trajectories. We use
16 NVIDIA H800 GPUs with 141\,GB of memory each, BF16 precision, a maximum
sequence length of 14K, a per-GPU batch size of 1, and eight
gradient-accumulation steps, yielding a global batch size of 128. Optimization
uses fused AdamW with \(\beta_1=0.9\), \(\beta_2=0.95\), weight decay \(0.1\),
and gradient clipping at \(1.0\). The learning rate is warmed up over the
first \(3\%\) of training to \(5\times10^{-6}\) and then decayed to
\(5\times10^{-7}\) with a cosine schedule.

Second, we initialize from the SFT checkpoint at step 16,200 and perform
400 steps of Group Relative Policy Optimization (GRPO)
\citep{shao2024deepseekmath} on 32 interactive OpenHA training tasks in
Minecraft, spanning navigation, world interaction, and GUI manipulation.
Each task is sampled with a group of eight online trajectories. GRPO uses
Adam with \(\beta_1=0.9\), \(\beta_2=0.95\), a constant learning rate of
\(1\times10^{-7}\), weight decay \(0.01\), gradient clipping at \(1.0\),
and a clipping ratio of \(0.2\). Training uses BF16, a policy micro-batch
size of 1, and one optimization epoch per rollout batch. We apply no
additional KL reward penalty. During rollout collection, actions are sampled
with temperature \(1.0\) and top-\(p=1.0\), and the policy receives a binary
task-completion reward.

We use the GRPO checkpoint at step 400 as the frozen VLA in all harness
experiments. Under the \texttt{legacy} protocol, each prediction conditions
only on the current observation and task instruction. Under \texttt{mtv1}
(\texttt{minecraft\_v1}), the model additionally receives up to ten preceding
observation--action turns. Both protocols predict four consecutive low-level
actions per decision, each executed for \(50\,\mathrm{ms}\), producing a
\(200\,\mathrm{ms}\) action chunk. The primary evaluation uses
\texttt{legacy} with temperature \(0.7\); auxiliary protocol comparisons use
temperature \(0\). The VLA parameters remain frozen throughout harness
evolution.

\subsection{Evaluation Setup}
\label{eval_setup}

We evaluate on 149 OpenHA tasks in Minecraft: 50 mining tasks, 49 combat
tasks, and 50 crafting tasks. Each episode permits up to 100 VLA decisions
and therefore up to 400 temporally dependent low-level actions. We use
\emph{long-horizon} to describe this multi-turn closed-loop dependency,
which may involve locating, approaching, aligning with, and interacting with
a target, followed by progress verification and failure recovery.

Evaluation uses eight NVIDIA H800 GPUs and eight parallel environment
workers. Each worker hosts one frozen policy instance and interacts with an
independent MineStudio environment. Task success is determined by the binary
completion signal provided by OpenHA. All methods within a comparison use the
same VLA checkpoint, task configurations, prompting protocol, temperature,
and interaction budget.

Under the primary \texttt{legacy}, temperature-\(0.7\) setting, Bare solves
39 of the 149 tasks. Its remaining 110 failures form the
\emph{Bare-failure adaptation subset}. This subset is defined operationally
by the frozen Bare policy and is not an independently curated difficulty
split.

Candidate and incumbent harnesses are compared using the same task and search
seed. Accepted candidates are subsequently evaluated using a disjoint
holdout seed. These evaluation seeds control harness-side stochasticity,
while the Minecraft world configuration is specified by the OpenHA task.
Paired conditions share the same VLA parameters, decoding settings,
episodic-memory snapshot, and probing-policy state. Persistent-memory updates
and probing-policy writeback are disabled during paired and holdout
evaluation.

After adaptation, we freeze the selected harness and probing statistics and
replay all 149 tasks without further search or writeback. Because this replay
includes tasks from the adaptation subset, it measures task coverage rather
than generalization to an unseen task split. The disjoint-seed evaluation
separately measures cross-seed robustness.

\subsection{Researcher and Sandbox Verifier.}
We use the API model identifier \texttt{gemini-2.5-flash}
\citep{comanici2025gemini} with temperature \(0\), visual input enabled, and
at most four trajectory frames per proposal. Its output is restricted to a
structured patch over the exposed harness configuration. Auto-K and
\method{} use the same proposer model, prompt format, temperature, and frame
budget.

All candidate patches are empirically evaluated in the MineStudio
sandbox \citep{ju2026mineexplorer}. The candidate and incumbent share the
same task, frozen VLA, decoding configuration, evaluation seed, and initial
external state. Persistent memory and probing-policy writeback are disabled
during comparison. The sandbox outcome, rather than the proposer itself,
determines whether a patch is accepted.

\subsection{Knowledge Search and Adaptive Probing Diagnostics}
\label{sec:knowledge_probe_diagnostics}

We further analyze the 110 tasks in the Bare-failure adaptation subset. This
analysis separates successes observed during Knowledge Search from those
obtained in probing-enabled replay. Because it aggregates multiple
intermediate harnesses, it is a search-time diagnostic rather than a
decomposition of the final frozen-harness coverage in
Table~\ref{tab:openha_results}.

\begin{table}[htbp]
\centering
\caption{
Search-time diagnostics on the 110 OpenHA tasks failed by Bare under the
primary evaluation setting. The union counts distinct tasks solved by either
Knowledge Search or probing-enabled replay.
}
\label{tab:knowledge_probe_diagnostics}
\resizebox{0.65\columnwidth}{!}{
\begin{tabular}{l|c|ccc}
\hline
Statistic & Overall & Mine & Kill & Craft \\
\hline
Knowledge Search Best
& 28/110 & 8 & 14 & 6 \\
Probing-enabled Replay
& 17/110 & 4 & 9 & 4 \\
\rowcolor{blue!8}
Search Best \(\cup\) Probing Replay
& 31/110 & 9 & 15 & 7 \\
\hline
ACCEPT
& 14 & 4 & 5 & 5 \\
HO-win
& 5 & 1 & 2 & 2 \\
Strict Probe-only
& 3 & 1 & 1 & 1 \\
\hline
\end{tabular}
}
\end{table}

Knowledge Search Best counts a task as solved if its best validated candidate
succeeds during search; it is therefore an aggregate search-time statistic,
not the performance of one deployable harness. It solves 28 tasks, while
probing-enabled replay succeeds on 17. Fourteen tasks overlap, giving
\begin{equation}
\left|
\mathcal{S}_{\mathrm{search}}
\cup
\mathcal{S}_{\mathrm{probe}}
\right|
=
28+17-14
=
31.
\end{equation}
Thus, the 17 replay successes are not 17 additional gains. Instead, three are
strict probe-only successes: one mining, one combat, and one crafting task
that remain unsolved by Knowledge Search but succeed during probing-enabled
replay.

The 31-task diagnostic union differs from the 31 rescued tasks in
Table~\ref{tab:openha_results}. The former combines successes from different
intermediate harnesses during search and probing replay, whereas the latter
is measured from the final evaluation used to report overall task coverage.
The two quantities therefore describe different stages of the evaluation and
should not be interpreted as identical statistics.

The combined diagnostic procedures solve 15 of the 31 Bare-failed combat
tasks (\(48.4\%\)), compared with 9 of 30 mining tasks (\(30.0\%\)) and
7 of 49 crafting tasks (\(14.3\%\)). Combat failures frequently involve
target loss or stalled engagement, where approaching, attacking, and
re-centering provide useful corrective evidence. Crafting remains more
difficult because it often requires precise multi-step GUI interaction.

Knowledge Search produces accepted patches for 14 distinct tasks, five of
which obtain an HO-win on a disjoint seed, corresponding to an empirical
holdout rate of \(35.7\%\). Because both search and holdout evaluation use a
limited number of seeds, these results provide preliminary cross-seed evidence
rather than a statistically conclusive generalization estimate.

\subsection{Sequential Cross-Task Knowledge Reuse}
\label{sec:sequential-knowledge-reuse}

During harness adaptation, tasks are processed sequentially. After each task,
validated recovery knowledge and contextual-bandit statistics are written to
the shared harness state and become available to subsequent tasks. We analyze
all 110 tasks in the Bare-failure adaptation subset from the
temperature-\(0.7\) run to verify that this information is used during later
execution.

We define a \emph{prior-knowledge selection} as an event in which the current
task selects a promoted recovery strategy whose provenance includes at least
one preceding task. A \emph{seed-harness success} is a task completed by the
inherited harness during its initial evaluation, before any task-specific
candidate is proposed or accepted. Each initial evaluation contains exactly
one episode, so these successes are not selected from repeated task restarts.

\begin{table}[t]
\centering
\small
\caption{
Sequential reuse of recovery knowledge in the temperature-\(0.7\)
adaptation run, computed from initial seed-rollout metrics before
task-specific proposal and acceptance.
}
\label{tab:sequential-knowledge-reuse}
\begin{tabular}{lr}
\toprule
Statistic & Count \\
\midrule
Adaptation tasks & 110 \\
Tasks using prior-task knowledge & 107 \\
Prior-knowledge selection events & 840 \\
Distinct prior-skill--target relations & 156 \\
Cross-family reuse relations & 101 \\
Single-rollout seed-harness successes & 12 \\
Successes without local ACCEPT & 12 \\
Successes explicitly using prior knowledge & 9 \\
\bottomrule
\end{tabular}
\end{table}

As shown in Table~\ref{tab:sequential-knowledge-reuse}, 107 of the 110
adaptation tasks select knowledge promoted from earlier tasks, yielding 840
selection events. The inherited harness completes 12 tasks before
task-specific search, and nine of these successful trajectories explicitly
use prior-task knowledge.

These results provide operational evidence of sequential cross-task reuse,
including transfer across task families. However, the inherited harness also
contains updated prompts and contextual-bandit state, while decoding at
temperature \(0.7\) remains stochastic. We therefore interpret this analysis
as evidence that knowledge transfer occurs, rather than as a causal estimate
of the improvement attributable solely to transferred knowledge.

\section{Multi-Seed Robustness on Bare-Failure Tasks}
\label{app:multiseed_robustness}

We evaluate cross-seed robustness on the 110 tasks that Bare fails under
the legacy protocol with temperature $0.7$. This diagnostic subset
contains 30 mining, 31 combat, and 49 crafting tasks. All conditions use
the same frozen VLA checkpoint. For the cross-seed transfer evaluation,
Auto-K executes its frozen harness and transferable knowledge bank
without test-time writeback, whereas \method{} retains the
knowledge--probing feedback loop. Table~\ref{tab:openha_hard110_per_seed}
reports the complete results for seeds $\{101,202,303\}$.

\paragraph{Consistent improvements across seeds.}
Auto-K solves 24, 8, and 15 tasks under seeds 101, 202, and 303,
respectively, while \method{} solves 31, 30, and 25. Thus,
\method{} improves coverage by 7, 22, and 10 tasks, with a positive
gain under every evaluated seed. Averaged across seeds, the number of
rescued tasks increases from
$15.7_{\scriptscriptstyle\pm 8.0}$ to
$28.7_{\scriptscriptstyle\pm 3.2}$ out of 110. Equivalently, the mean
success rate improves from
$14.2_{\scriptscriptstyle\pm 7.3}\%$ to
$26.1_{\scriptscriptstyle\pm 2.9}\%$.
The lower standard deviation additionally indicates that \method{} is
less sensitive to the evaluation seed than Auto-K.

\paragraph{Improvements across task families.}
Aggregated over the three seeds, Auto-K solves 47 of 330 task--seed
instances, whereas \method{} solves 86. The improvement is distributed
across all three task families. Mining success increases from 14 to 25
instances, combat success from 23 to 41, and crafting success from 10
to 20. The gain therefore cannot be attributed solely to one task
family or to an unusually favorable environment configuration.

\paragraph{Task-paired evidence.}
A task-paired analysis provides stronger evidence than comparing only
aggregate success rates. Under seed 202, \method{} exclusively solves
24 tasks that Auto-K fails, whereas Auto-K exclusively solves only two
tasks. Under seed 303, the corresponding counts are 12 and two.
Pooling these two rerun seeds gives 36 \method{}-only successes versus
four Auto-K-only successes. An exact McNemar test rejects symmetric
random outcome fluctuations
($p=1.86\times10^{-7}$). Thus, the observed difference is unlikely to
arise from a small number of stochastic success flips.

\paragraph{Contribution of adaptive probing.}
Across the three seeds, \method{} produces 3, 3, and 1 strict
probe-only recoveries and obtains 17, 11, and 16 successful
probing-enabled replays. On average, this corresponds to
$2.3_{\scriptscriptstyle\pm 1.2}$ strict probe-only recoveries and
$14.7_{\scriptscriptstyle\pm 3.2}$ replay successes per seed.
Meanwhile, Knowledge Search produces
$14.0_{\scriptscriptstyle\pm 3.0}$ accepted task-level patches, of
which $6.7_{\scriptscriptstyle\pm 2.1}$ generalize to the holdout
evaluation. These observations suggest that the gain is not explained
merely by generating more patches. Instead, probe feedback changes
which failure contexts are explored and which recovery knowledge is
selected and retained.

\paragraph{Interpretation.}
The consistent improvement across all seeds, the reduction in
cross-seed variance, and the strongly asymmetric task-level success
flips jointly support the system-level benefit of knowledge--probing
co-evolution. This experiment evaluates the complete feedback loop
rather than attributing the gain to either direction individually.

\vspace{-.3cm}
\begin{table*}[htbp]
\centering
\caption{
Per-seed results on the 110 tasks failed by Bare under the legacy
protocol with temperature $0.7$. The subset contains 30 mining,
31 combat, and 49 crafting tasks.
}
\label{tab:openha_hard110_per_seed}
\resizebox{\textwidth}{!}{
\begin{tabular}{l|c|c|ccc|cc|cc}
\hline
Method
& Seed
& Success
& \multicolumn{3}{c|}{Task success}
& \multicolumn{2}{c|}{Knowledge search}
& \multicolumn{2}{c}{Adaptive probing} \\
\cline{4-6}\cline{7-8}\cline{9-10}
&
&
& Mine & Kill & Craft
& ACCEPT & HO-win
& P-only & Replay succ. \\
\hline

Auto-K
& 101
& 24/110 (21.8\%)
& 5/30 & 12/31 & 7/49
& 14 & 5
& N/A & N/A \\

\rowcolor{blue!8}
\method{}
& 101
& \textbf{31/110 (28.2\%)}
& \textbf{9/30} & \textbf{15/31} & \textbf{7/49}
& 14 & 5
& \textbf{3} & \textbf{17} \\

\hline

Auto-K
& 202
& 8/110 (7.3\%)
& 5/30 & 2/31 & 1/49
& -- & --
& N/A & N/A \\

\rowcolor{blue!8}
\method{}
& 202
& \textbf{30/110 (27.3\%)}
& \textbf{9/30} & \textbf{14/31} & \textbf{7/49}
& 17 & 9
& \textbf{3} & \textbf{11} \\

\hline

Auto-K
& 303
& 15/110 (13.6\%)
& 4/30 & 9/31 & 2/49
& -- & --
& N/A & N/A \\

\rowcolor{blue!8}
\method{}
& 303
& \textbf{25/110 (22.7\%)}
& \textbf{7/30} & \textbf{12/31} & \textbf{6/49}
& 11 & 6
& \textbf{1} & \textbf{16} \\

\hline

Auto-K
& Mean
& \(15.7_{\scriptscriptstyle\pm 8.0}/110
   \;(14.2_{\scriptscriptstyle\pm 7.3}\%)\)
& \(4.7_{\scriptscriptstyle\pm 0.6}/30\)
& \(7.7_{\scriptscriptstyle\pm 5.1}/31\)
& \(3.3_{\scriptscriptstyle\pm 3.2}/49\)
& -- & --
& N/A & N/A \\

\rowcolor{blue!8}
\method{}
& Mean
& \(\mathbf{28.7}_{\scriptscriptstyle\pm 3.2}/110
   \;(\mathbf{26.1}_{\scriptscriptstyle\pm 2.9}\%)\)
& \(\mathbf{8.3}_{\scriptscriptstyle\pm 1.2}/30\)
& \(\mathbf{13.7}_{\scriptscriptstyle\pm 1.5}/31\)
& \(\mathbf{6.7}_{\scriptscriptstyle\pm 0.6}/49\)
& \(14.0_{\scriptscriptstyle\pm 3.0}\)
& \(6.7_{\scriptscriptstyle\pm 2.1}\)
& \(\mathbf{2.3}_{\scriptscriptstyle\pm 1.2}\)
& \(\mathbf{14.7}_{\scriptscriptstyle\pm 3.2}\) \\

\hline
\end{tabular}
}

\vspace{1mm}
\begin{minipage}{0.99\textwidth}
\footnotesize
The Mean rows report the mean and sample standard deviation over seeds
$\{101,202,303\}$, with the standard deviation shown as a lower-right
subscript. P-only denotes strict probe-only recoveries, for which
Knowledge Search fails but the probing-enabled replay succeeds.
Replay succ. counts all successful probing-enabled replays.
``N/A'' indicates that Auto-K does not contain the adaptive probing and
delayed-credit mechanism. A dash indicates that Knowledge Search was
disabled during the corresponding frozen Auto-K evaluation and therefore
ACCEPT and HO-win are not defined.
\end{minipage}
\end{table*}

\section{Static-Knowledge Diagnostic}
\label{sec:static_knowledge_diagnostic}

We conduct an additional diagnostic to examine whether injecting
a fixed recovery bank improves adaptive probing.
The diagnostic uses the 110 tasks failed by Bare under the
\texttt{legacy} protocol at temperature $0.7$, with the same frozen
VLA, evaluation seed 101, task list, base harness, and probe budget
of eight.
Each variant starts with an empty bandit state and performs one
execution rollout per task.
Knowledge content and harness patches remain fixed:
no proposer, Knowledge Search, \textsc{Accept}/\textsc{Revert},
holdout validation, or skill creation is enabled.
Environment feedback may update selection statistics, but cannot
create or revise knowledge.

\paragraph{Variants.}
All four variants use adaptive probing and update probe-selection
statistics from outcomes.
\emph{Probe-only} uses no knowledge bank.
\emph{Static K (no injection)} updates knowledge-selection
statistics but does not inject selected knowledge into the actor.
\emph{Static K (no K-update)} injects selected knowledge but
disables outcome-based updates to knowledge-selection statistics.
\emph{Static K (adaptive selection)} enables both knowledge
injection and outcome-based updates to its selection statistics.

\begin{table*}[t]
\centering
\caption{
Task success on the 110-task Bare-failure subset under
\texttt{legacy} at temperature $0.7$.
Panel (a) reports the complete search pipelines;
Panel (b) reports the fixed-bank execution diagnostic at seed 101.
}
\label{tab:knowledge_sandbox_diagnostic}
\begingroup
\small
\renewcommand{\arraystretch}{1.12}
\setlength{\tabcolsep}{4pt}
\begin{tabular*}{\textwidth}{
@{\extracolsep{\fill}}lcccc@{}
}
\hline
Method / Variant
& Success
& Mine
& Kill
& Craft \\
\hline
\multicolumn{5}{c}{
\emph{(a) Full Sandbox-mediated knowledge evolution}
} \\
\hline

Auto-K
& 24/110 (21.8\%)
& 5/30
& 12/31
& \textbf{7/49} \\

\rowcolor{blue!8}
\method{}
& \textbf{31/110 (28.2\%)}
& \textbf{9/30}
& \textbf{15/31}
& \textbf{7/49} \\

\hline
\multicolumn{5}{c}{
\emph{(b) Static-knowledge execution diagnostic}
} \\
\hline

Probe-only
& \textbf{15/110 (13.6\%)}
& 5/30
& \textbf{6/31}
& \textbf{4/49} \\

Static K (no injection)
& 9/110 (8.2\%)
& 2/30
& \textbf{6/31}
& 1/49 \\

Static K (no K-update)
& 13/110 (11.8\%)
& 4/30
& 5/31
& \textbf{4/49} \\

Static K (adaptive selection)
& 13/110 (11.8\%)
& \textbf{6/30}
& 5/31
& 2/49 \\

\hline
\end{tabular*}
\endgroup

\vspace{1mm}
\begin{minipage}{0.99\textwidth}
\footnotesize
Panel (a) is the hard-subset view of
Table~\ref{tab:openha_results}: including the 39 Bare successes
gives 63/149 for Auto-K and 70/149 for \method{}.
Panel (b) uses one execution rollout per task with fixed knowledge
content and harness patches.
Results are comparable within each panel; cross-panel differences
are not controlled effect estimates.
Bold denotes the highest coverage within each panel.
\end{minipage}
\end{table*}

\begin{table*}[t]
\centering
\caption{
Process statistics for the static-knowledge diagnostic.
Positive-utility K is the percentage of knowledge selections
with positive immediate utility.
}
\label{tab:static_knowledge_process}
\begingroup
\small
\renewcommand{\arraystretch}{1.12}
\setlength{\tabcolsep}{3pt}
\begin{tabular*}{\textwidth}{
@{\extracolsep{\fill}}lccccc@{}
}
\hline
Variant
& \shortstack{Probes\\per task}
& \shortstack{Steps to\\recovery}
& \shortstack{Useful\\probes (\%)}
& \shortstack{K\\selections}
& \shortstack{Positive-utility\\K (\%)} \\
\hline

Probe-only
& 8.00 & 59.4 & 42.7 & -- & -- \\

Static K (no injection)
& 7.98 & 50.8 & 41.3 & 1,463 & 0.07 \\

Static K (no K-update)
& 7.98 & 61.8 & 39.6 & 1,454 & 0.00 \\

Static K (adaptive selection)
& 7.93 & 51.5 & 41.7 & 1,428 & 0.14 \\

\hline
\end{tabular*}
\endgroup
\end{table*}

\paragraph{Results.}
None of the tested static-bank variants outperforms Probe-only
in overall coverage
(Table~\ref{tab:knowledge_sandbox_diagnostic}).
Adaptive selection over the fixed bank solves 13 tasks,
compared with 15 for Probe-only.
Both variants solve seven tasks; Probe-only uniquely solves eight,
and adaptive static-K selection uniquely solves six.
The paired difference is not statistically significant
(two-sided exact McNemar test, $p=0.79$).
Their observed probe counts and useful-probe fractions are also
similar (Table~\ref{tab:static_knowledge_process}).
Thus, adding this fixed bank provides no observed coverage
benefit under the evaluated setting.

Adaptive static-K selection makes 1,428 knowledge selections,
but only 0.14\% have positive immediate utility.
This metric captures immediate feedback rather than the full
downstream contribution of knowledge.
The outcomes also vary across task families:
mining coverage increases from five tasks to six, while crafting
coverage decreases from four to two.
The bank primarily contains generic visual recovery strategies,
including target reacquisition, re-centering, and escaping walls.
These strategies may offer limited help for crafting-interface
or prerequisite failures, suggesting that recovery depends on
the relevance of knowledge to the encountered failure context.

\paragraph{Why Sandbox-mediated knowledge--probing evolution matters.}
Sandbox provides the execution evidence that connects recovery
knowledge with adaptive probing.
Knowledge directs probes toward unresolved failure hypotheses,
while probe outcomes reveal whether recovery strategies are
executable by the frozen actor.
This evidence informs knowledge selection and the proposal
and validation of harness updates.
Validated recovery experience can then guide subsequent probing
and be reused across tasks, enabling knowledge and probing
to adapt together.

The diagnostic highlights the distinction between updating
selection statistics and revising the knowledge being selected.
All diagnostic variants interact with the environment, but
their feedback cannot improve the bank's content.
In the complete framework, Sandbox feedback additionally drives
knowledge acquisition, harness revision, and validation.
The main results demonstrate the benefit of this overall design:
coverage increases from 50/149 for Fixed to 63/149 for Auto-K
and 70/149 for \method{}.
The advantage over Auto-K also holds across all three evaluated
seeds, increasing the mean rescue count from
$15.7 \pm 8.0$ to $28.7 \pm 3.2$ on the same 110-task subset.
Together, these findings support Sandbox-mediated
knowledge--probing co-evolution as an effective approach to
harness adaptation, with Sandbox serving as the shared
execution and validation interface for accumulating
reusable recovery experience.

\paragraph{Scope.}
The static-bank diagnostic evaluates one bank and one seed under
a restricted execution protocol.
It retains environment interaction and differs from the complete
search-and-replay pipeline.
Its results therefore provide complementary system-level evidence,
without isolating the causal contribution of Sandbox interaction
or individual adaptation components.

\section{Harness Search Implementation Details}
\label{sec:harness_search_details}

\paragraph{Researcher and search budget.}
We use Gemini-2.5-Flash \citep{comanici2025gemini} at temperature \(0\)
as both the visual failure-attribution researcher and harness proposer.
It receives the incumbent harness, rollout statistics, failure and probe
records, relevant recovery knowledge, previous proposal outcomes, and up
to four frames sampled from the failed trajectory. Each task uses three
sequential proposal rounds with at most one candidate per round. Invalid
outputs are repaired for up to three researcher calls and discarded if
they remain invalid. An accepted candidate becomes the incumbent for the
next round, whereas a rejected candidate leaves the incumbent unchanged.
Auto-K and \method{} use the same researcher and proposal budget.

\paragraph{Editable harness configuration.}
The researcher may modify the task, decomposition, and recovery prompts;
the runtime task instruction; and selected probe, memory, verification,
recovery, and runtime settings. Every candidate must contain at least one
prompt or instruction edit, and parameter-only patches are rejected before
execution. The researcher cannot modify the VLA checkpoint or parameters,
the low-level action space, task definition, MineExplorer environment, or
evaluation protocol.

\paragraph{Failure contexts and runtime signals.}
The action-probing context combines a coarse task family with the event
that activates probing, producing contexts such as
\texttt{mine:stall}, \texttt{combat:fail}, and
\texttt{craft:stall}. The knowledge-probing context additionally includes
failure attributes such as \texttt{zero\_progress},
\texttt{orientation}, \texttt{target\_loss}, and
\texttt{stall\_heavy}. These attributes are obtained from rollout
statistics and visual attribution over the sampled trajectory frames.

Progress and reward are obtained from environment feedback, while execution
failure is determined from the runtime status. Novelty is computed as one
minus the maximum similarity between the current state--action pair and
previously stored experience:
\begin{equation}
n_t
=
1-
\max_{(z_i,e_i)\in\mathcal{M}}
\frac{1}{2}
\left[
\cos(z_t,z_i)+\cos(e(a_t),e_i)
\right],
\label{eq:probe_novelty}
\end{equation}
where \(z_t\) denotes the visual-state representation and \(e(a_t)\) the
action embedding. We set \(n_t=1\) when no previous experience is available.

\paragraph{Paired acceptance.}
For an episode executed by harness \(H\), the acceptance score is
\begin{equation}
J(H)
=
\mathbb{I}[\mathrm{success}]
+0.005R_{\mathrm{env}}
-0.2\mathbb{I}[\mathrm{cascade}],
\label{eq:harness_acceptance_score}
\end{equation}
where \(R_{\mathrm{env}}\) is the raw cumulative environment reward without
additional normalization. Progress, interaction cost, and step efficiency
are recorded as diagnostic statistics and may be supplied to the researcher
for failure analysis and proposal generation, but they do not enter the
ACCEPT/REVERT decision. Since the environment reward is not assumed to have
a globally fixed range, we do not interpret Equation~\eqref{eq:harness_acceptance_score}
as a strictly lexicographic preference for success.

The candidate and incumbent are evaluated on the same task using one search
seed, identical decoding parameters, and the same initial memory and
probing-policy state. A candidate \(H'\) is accepted when
\begin{equation}
J(H')-J(H)\geq 0.01.
\label{eq:harness_acceptance_rule}
\end{equation}
Rejected candidates leave the incumbent unchanged. Accepted candidates are
additionally compared with the seed harness using another disjoint holdout
seed. Holdout results do not alter the original acceptance decision.
Persistent memory and probing-policy writeback are disabled during both
paired and holdout evaluation.

\paragraph{Example harness patch.}
The following shortened example shows an actual candidate generated for
the sand-mining task:

\begin{footnotesize}
\begin{verbatim}
{
  "diagnosis":
    "The agent stalls while sand is visible.",
  "why":
    "The instruction does not specify target
     centering and persistent interaction.",
  "proposal":
    "Locate sand, center it, and break it.",
  "edits": [
    {
      "path": ["runtime", "instruction"],
      "value":
        "Locate sand, center it, then break it."
    },
    {
      "path": ["recovery_prompt"],
      "value":
        "If sand is not centered, re-orient and
         retry. <actions> mouseMove(0,80);
         mouseMove(0,40); mouseClick(left);
         no_op </actions>"
    }
  ]
}
\end{verbatim}
\end{footnotesize}

The patch modifies only exposed harness fields and leaves the VLA frozen.
It is retained only if its paired execution improves over the incumbent
according to Equation~\ref{eq:harness_acceptance_rule}.

\section{Harness Evolution and Evaluation}
\label{sec:harness_setup}

\paragraph{Initial Gaming Harness.}
We apply \method{} to the 110 tasks in the Bare-failure adaptation subset.
For every task, the actor is fixed to the GRPO checkpoint at step 400 and
uses the primary \texttt{legacy} protocol with sampling temperature \(0.7\)
and four low-level actions per prediction. The VLA parameters are never
updated during harness evolution.

At evolution round \(k\), the external harness \(H_k\) maintains action
probes, task knowledge, transferable recovery memory, probing statistics,
and action-level middleware. The middleware operates in the
latency-sensitive execution loop and can adjust camera motion, sustain or
interrupt an action, select an inventory item, invoke recovery behavior, or
verify task completion without modifying the underlying VLA.

\paragraph{Adaptive Gaming Probing.}
When execution stalls, repeats an ineffective action, loses its target,
selects an inappropriate tool, or fails to recognize completion, the harness
activates adaptive probing. Given the current task and failure context, it
constructs or specializes applicable action probes from the fixed low-level
action space. These probes test execution-relevant hypotheses, including
whether the target is aligned, within interaction range, or requires
persistent interaction. Their visual transitions, environment feedback, and
progress signals refine the inferred failure context.

The knowledge-probing policy then selects a stored recovery strategy as a
hypothesis, specializes it to the current task, and tests it through
subsequent interaction. Probe choices receive immediate credit from progress,
reward, novelty, and failure resolution, together with delayed credit from
episode success. Effective probes and knowledge are preferred in related
contexts through the updated probing policies. Consequently, probing adapts
both the interventions considered for a failure and their context-conditioned
selection without modifying the underlying VLA.

\begin{figure*}[t]
\centering
\includegraphics[width=\textwidth]{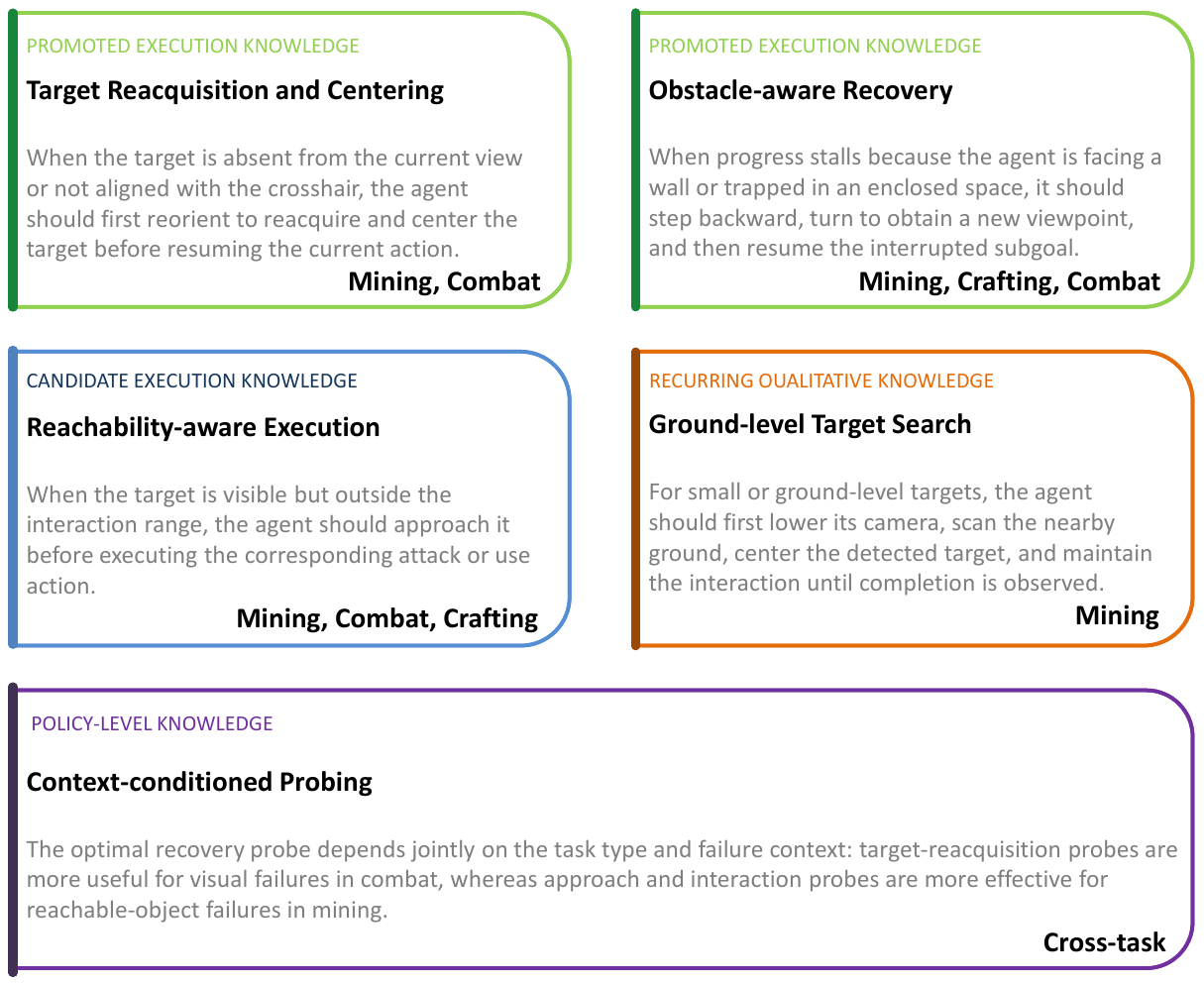}
\vspace{-.2cm}
\caption{
Representative execution knowledge acquired by \method{}. Promoted recovery
strategies (green), candidate knowledge (blue), recurring qualitative
patterns (orange), and context-conditioned probing knowledge (purple)
connect task and failure contexts to reusable interventions across mining,
combat, and crafting.
}
\label{fig:knowledge}
\end{figure*}

\begin{figure*}[t]
\centering
\includegraphics[width=\textwidth]{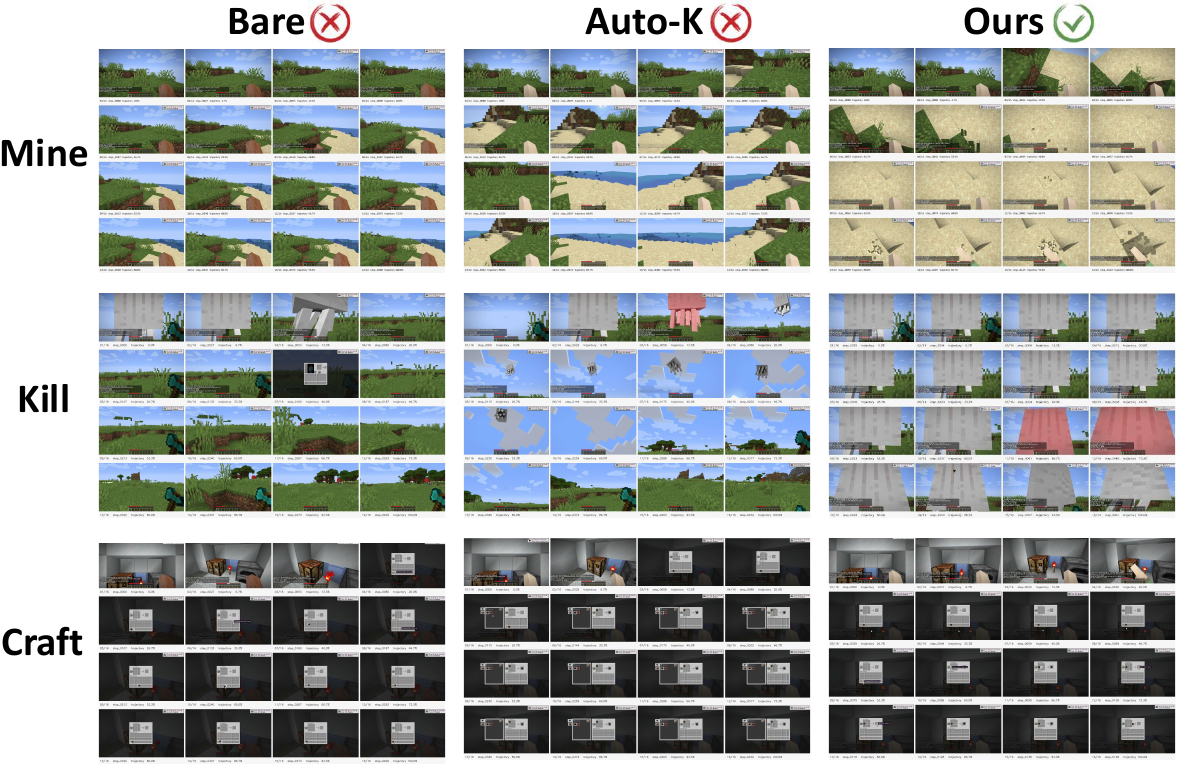}
\vspace{-.2cm}
\caption{
Qualitative comparisons on mining, combat, and crafting tasks using the same
frozen VLA. In these examples, Bare and Auto-K fail, while \method{} uses
adaptive probing to identify and test applicable recovery knowledge before
completing the task.
}
\label{fig:examples}
\end{figure*}

\begin{table*}[t]
\centering
\small
\caption{
Low-level Minecraft action space used by the frozen VLA. Multiple commands
may be composed within the same action step.
}
\label{tab:minecraft-action-space}
\label{action}

\begin{tabularx}{\textwidth}{
  @{}
  p{0.25\textwidth}
  X
  @{}
}
\toprule
\textbf{Action} & \textbf{Execution semantics} \\
\midrule

\texttt{mouseMove(dx,dy)}
&
Move the camera horizontally and vertically. The values \(dx\) and \(dy\)
determine the direction and magnitude of camera rotation.
\\[3pt]

\texttt{mouseClick(left)}
&
Attack the targeted entity or block. In an inventory GUI, pick up, place,
or collect an item stack.
\\[3pt]

\texttt{mouseClick(right)}
&
Use the selected item, place a block, interact with the targeted object,
or manipulate an item in an inventory GUI.
\\[3pt]

\texttt{mouseClick(middle)}
&
Execute a middle-button mouse interaction.
\\[3pt]

\texttt{keyPress(w)}
&
Move forward.
\\[3pt]

\texttt{keyPress(s)}
&
Move backward.
\\[3pt]

\texttt{keyPress(a)}
&
Strafe left.
\\[3pt]

\texttt{keyPress(d)}
&
Strafe right.
\\[3pt]

\texttt{keyPress(space)}
&
Jump.
\\[3pt]

\texttt{keyPress(e)}
&
Open or close the inventory and the \(2\times2\) crafting grid.
\\[3pt]

\texttt{keyPress(ESC)}
&
Close the currently opened inventory or GUI.
\\[3pt]

\texttt{keyPress(q)}
&
Drop one item from the selected item stack.
\\[3pt]

\texttt{keyPress(1--9)}
&
Select the corresponding hotbar slot.
\\[3pt]

\texttt{keyPress(left.ctrl)}
&
Sprint during movement. When combined with \texttt{q}, drop the entire
selected stack.
\\[3pt]

\texttt{keyPress(left.shift)}
&
Move carefully or perform a modified inventory operation, such as
transferring a stack between the inventory and hotbar.
\\[3pt]

\texttt{no\_op}
&
Wait for one \(50\,\mathrm{ms}\) action step without interacting with the
environment.
\\

\bottomrule
\end{tabularx}

\vspace{4pt}

\parbox{0.96\textwidth}{
\textbf{Action composition.}
Commands can be composed within one action step, for example
\texttt{mouseMove(10,10) mouseClick(left) keyPress(w)}.
Each VLA prediction contains exactly four semicolon-separated action steps:
\texttt{<actions> \(a_1;a_2;a_3;a_4\) </actions>}.
Each step lasts \(50\,\mathrm{ms}\), yielding a \(200\,\mathrm{ms}\)
execution horizon per prediction.
}
\end{table*}

\begin{figure*}[t]
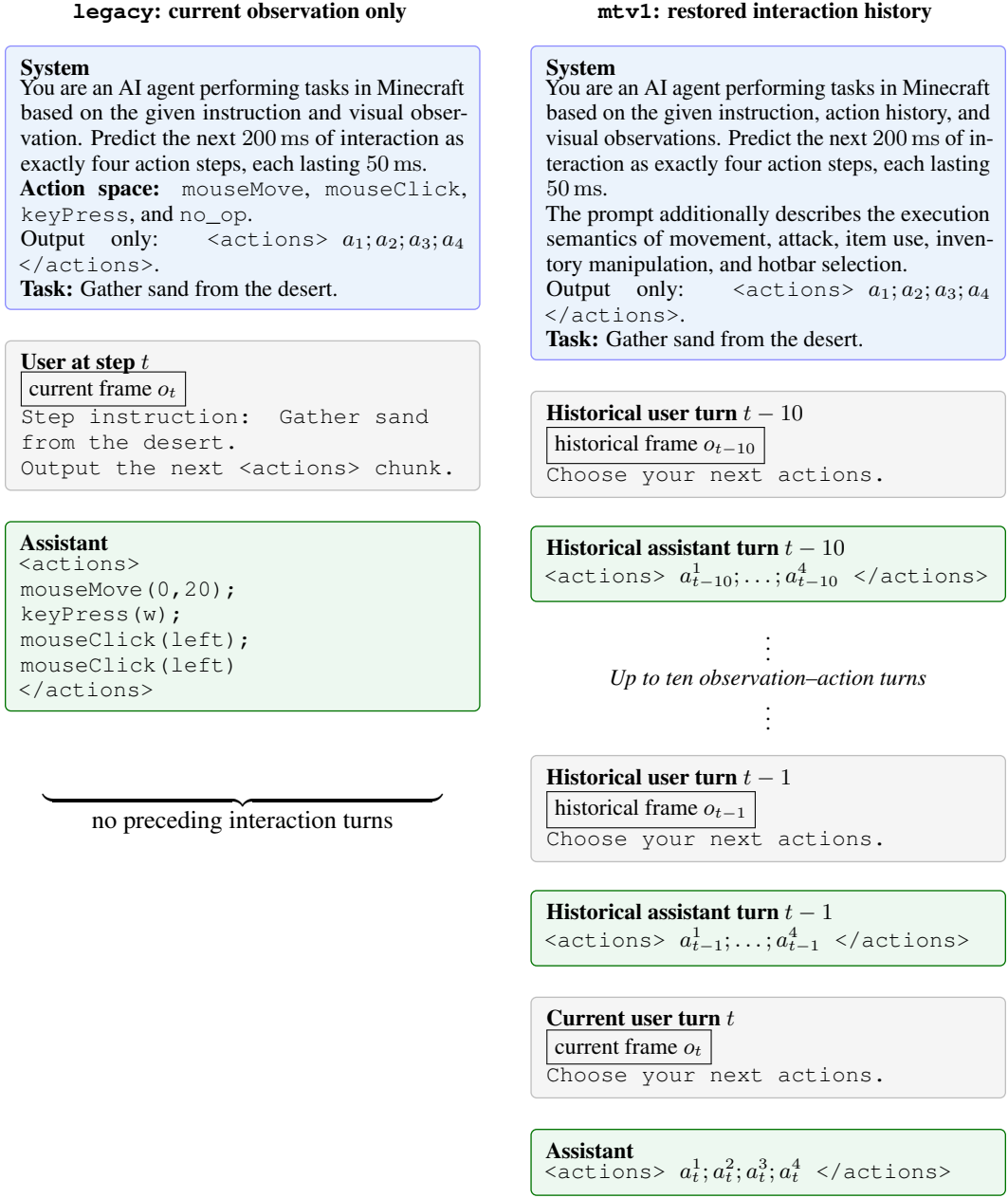

\centering
\small

\begin{minipage}[t]{0.475\textwidth}
\centering
\textbf{\texttt{legacy}: current observation only}
\vspace{3pt}

\begin{systemmessage}
\textbf{System}\\[-2pt]
You are an AI agent performing tasks in Minecraft based on the given
instruction and visual observation. Predict the next \(200\,\mathrm{ms}\)
of interaction as exactly four action steps, each lasting
\(50\,\mathrm{ms}\).

\textbf{Action space:}
\texttt{mouseMove}, \texttt{mouseClick},
\texttt{keyPress}, and \texttt{no\_op}.

Output only:
\texttt{<actions> \(a_1;a_2;a_3;a_4\) </actions>}.

\textbf{Task:} Gather sand from the desert.
\end{systemmessage}

\vspace{2pt}

\begin{usermessage}
\textbf{User at step \(t\)}\\[-2pt]
\(\boxed{\text{current frame }o_t}\)

\texttt{Step instruction: Gather sand from the desert.}\\
\texttt{Output the next <actions> chunk.}
\end{usermessage}

\vspace{2pt}

\begin{assistantmessage}
\textbf{Assistant}\\[-2pt]
\texttt{<actions>}\\
\texttt{mouseMove(0,20);}\\
\texttt{keyPress(w);}\\
\texttt{mouseClick(left);}\\
\texttt{mouseClick(left)}\\
\texttt{</actions>}
\end{assistantmessage}

\vspace{3pt}
\[
\displaystyle
\underbrace{\hspace{0.84\linewidth}}_{
  \text{\normalsize no preceding interaction turns}
}
\]

\end{minipage}
\hfill
\begin{minipage}[t]{0.475\textwidth}
\centering
\textbf{\texttt{mtv1}: restored interaction history}
\vspace{3pt}

\begin{systemmessage}
\textbf{System}\\[-2pt]
You are an AI agent performing tasks in Minecraft based on the given
instruction, action history, and visual observations. Predict the next
\(200\,\mathrm{ms}\) of interaction as exactly four action steps, each
lasting \(50\,\mathrm{ms}\).

The prompt additionally describes the execution semantics of movement,
attack, item use, inventory manipulation, and hotbar selection.

Output only:
\texttt{<actions> \(a_1;a_2;a_3;a_4\) </actions>}.

\textbf{Task:} Gather sand from the desert.
\end{systemmessage}

\vspace{2pt}

\begin{usermessage}
\textbf{Historical user turn \(t-10\)}\\[1pt]
\(\boxed{\text{historical frame }o_{t-10}}\)

\texttt{Choose your next actions.}
\end{usermessage}

\vspace{1pt}

\begin{assistantmessage}
\textbf{Historical assistant turn \(t-10\)}\\[1pt]
\texttt{<actions> \(a_{t-10}^{1};\ldots;a_{t-10}^{4}\) </actions>}
\end{assistantmessage}

\vspace{-3pt}

\begin{center}
\(\vdots\)

\textit{Up to ten observation--action turns}

\(\vdots\)
\end{center}

\vspace{-3pt}

\begin{usermessage}
\textbf{Historical user turn \(t-1\)}\\[1pt]
\(\boxed{\text{historical frame }o_{t-1}}\)

\texttt{Choose your next actions.}
\end{usermessage}

\vspace{1pt}

\begin{assistantmessage}
\textbf{Historical assistant turn \(t-1\)}\\[1pt]
\texttt{<actions> \(a_{t-1}^{1};\ldots;a_{t-1}^{4}\) </actions>}
\end{assistantmessage}

\vspace{2pt}

\begin{usermessage}
\textbf{Current user turn \(t\)}\\[1pt]
\(\boxed{\text{current frame }o_t}\)

\texttt{Choose your next actions.}
\end{usermessage}

\vspace{2pt}

\begin{assistantmessage}
\textbf{Assistant}\\[-2pt]
\texttt{<actions> \(a_t^1;a_t^2;a_t^3;a_t^4\) </actions>}
\end{assistantmessage}

\end{minipage}

\caption{
Comparison of the two VLA prompting protocols. \texttt{legacy} performs
stateless prediction from the current observation and task instruction,
whereas \texttt{mtv1} restores up to ten preceding observation--action turns.
Both protocols predict four \(50\,\mathrm{ms}\) low-level actions per call.
System prompts are shortened for readability; complete action semantics are
given in Table~\ref{tab:minecraft-action-space}.
}
\label{fig:vla_prompt_protocols}
\end{figure*}

\paragraph{Evidence-Grounded Harness Proposal.}
The slow research loop converts the inferred failure mode, probe evidence,
and relevant recovery knowledge into a candidate harness
\(\widetilde{H}_k\). A proposal may introduce or revise camera alignment,
attack persistence, reachability handling, loop breaking, tool selection,
completion verification, or task-knowledge retrieval. Because the proposal
is conditioned on interaction evidence, it targets an observed failure
mechanism rather than relying only on the textual description of an
unsuccessful episode.

Each candidate edit is represented as a trigger--intervention unit. The
trigger specifies the task, visual, and execution conditions under which the
rule should activate, while the intervention specifies the corresponding
action correction or recovery procedure. This representation allows
validated knowledge to be retrieved and specialized for related tasks
without modifying the VLA.

\paragraph{Paired Proposal Validation.}
We compare every candidate \(\widetilde{H}_k\) with its incumbent \(H_k\)
through paired execution in the MineStudio sandbox. Both conditions
use the same task configuration, frozen VLA checkpoint, decoding settings,
interaction budget, and evaluation seed. Candidate selection prioritizes task
completion, with progress, reward, step efficiency, and absence of regression
as secondary criteria.

The two conditions start from isolated copies of the same episodic memory and
probing-policy state. Memory updates and probing-policy writeback are disabled
during evaluation, preventing execution order from affecting the comparison.
If the candidate does not improve over the incumbent, it is rejected and
\(H_{k+1}=H_k\). Otherwise, it is accepted,
\(H_{k+1}=\widetilde{H}_k\), and its trigger, intervention, evidence, and
outcome form a candidate knowledge record.

\paragraph{Cross-Seed Verification and Knowledge Transfer.}
Accepted candidates are subsequently evaluated using a disjoint holdout seed.
The holdout result does not alter the original paired acceptance decision or
trigger further modification of that candidate; it provides separate evidence
of cross-seed transfer. We report an HO-win when the accepted candidate
outperforms its seed harness by the required margin on the holdout seed.

An accepted recovery is promoted to transferable knowledge when it obtains
holdout support or is successfully reused on another task. Promoted knowledge
enters the shared knowledge bank and becomes a candidate for future knowledge
probing. Separate learning rollouts, rather than paired or holdout evaluation,
update and persist the contextual probing statistics.

The resulting evolution cycle is
\[
\begin{aligned}
H_k
&\;\longrightarrow\;
\text{adaptive action and knowledge probing}
\;\longrightarrow\;
\text{failure attribution}
\;\longrightarrow\;
\widetilde{H}_k,
\\[3pt]
(H_k,\widetilde{H}_k)
&\;\longrightarrow\;
\text{paired sandbox validation}
\;\longrightarrow\;
H_{k+1}.
\end{aligned}
\]
A rejected proposal leaves the harness and persistent state unchanged.
An accepted proposal updates the harness, while separately validated
recoveries and probing statistics support subsequent tasks.

\paragraph{Final Harness Evaluation.}
After evolution, we freeze the selected harness, knowledge bank, and probing
statistics and replay the complete 149-task OpenHA evaluation set. The
39 tasks originally solved by Bare and the 110 tasks in the Bare-failure
adaptation subset are all included. No further harness search, memory
writeback, or probing-policy update is allowed, although the frozen probing
policy may still execute probes.

This final replay measures overall task coverage and checks whether the
evolved harness preserves tasks already solved by the frozen actor. Because
the replay includes tasks encountered during adaptation, it is reported as
task coverage rather than generalization to an unseen task split; disjoint
holdout seeds provide a separate measure of cross-seed robustness.

\section{Qualitative Examples}
\label{sec:probing_advantage}

The advantage of \method{} does not arise simply from executing additional
probe actions. It comes from using probe outcomes to connect failure
diagnosis, knowledge probing, harness adaptation, and cross-task knowledge
accumulation. We illustrate this distinction by comparing Bare, Auto-K, and
\method{} on a sand-mining task under the same frozen VLA.

\paragraph{Bare Execution.}
Bare executes the original task instruction without an external recovery
harness. When the policy loses the target, becomes trapped, or acts outside
the effective interaction range, it must recover using only the information
provided by the current prompting protocol. The failure is neither explicitly
diagnosed nor converted into reusable execution knowledge. Bare therefore
represents frozen execution without harness adaptation.

\paragraph{Auto-K.}
Auto-K searches for and validates recovery knowledge, making it stronger than
Bare. However, its probe actions are selected through predefined triggers or
coverage rules rather than a policy learned from probe outcomes. Consequently,
Auto-K may look up, look down, attack, or move forward, but the resulting
feedback does not update which action or knowledge probe should be preferred
under similar future failures.

This distinction is visible in the sand example. Auto-K executes eight probe
actions, but records no knowledge-selection or bandit update:
\begin{equation}
\texttt{knowledge\_events}=\varnothing,\qquad
\texttt{probe\_bandit}=\varnothing,\qquad
\texttt{knowledge\_bandit}=\varnothing.
\end{equation}
It continues to use the generic instruction
\emph{``Gather sand from the desert''} and a generic retry prompt. Its only
relevant prior is \texttt{reacquire\_and\_center}, inherited from an earlier
poppy task. Without failure-conditioned knowledge probing and task-specific
specialization, Auto-K does not complete the task within the 100-turn
interaction budget.

\paragraph{Closed-Loop Adaptation in \method{}.}
\method{} treats an action probe as an information-gathering intervention and
a knowledge probe as an executable recovery hypothesis. Action-probe outcomes
provide evidence about target loss, orientation error, insufficient
reachability, or stalled progress. Conditioned on this evidence, the
knowledge-probing policy selects a recovery strategy, which is specialized to
the current task and tested through subsequent interaction. Immediate utility
and delayed episode success then update the corresponding probing statistics.

The online adaptation loop is
\begin{equation}
\begin{aligned}
\text{action probe}
&\rightarrow
\text{failure-context inference}
\rightarrow
\text{knowledge probe}\\
&\rightarrow
\text{task-specific recovery}
\rightarrow
\text{interaction outcome}
\rightarrow
\text{credit update}.
\end{aligned}
\end{equation}
The resulting evidence may subsequently support a harness proposal, which is
accepted or rejected through paired sandbox validation. Persistent probing
statistics are updated during learning rollouts, not during paired or holdout
evaluation, where writeback is disabled.

For the sand task, \method{} loads
\texttt{backup\_and\_turn},
\texttt{reacquire\_and\_center}, and the candidate
\texttt{approach\_and\_act}. At decision turn 23, the context
\texttt{mine\_block:stall\_heavy} selects
\texttt{backup\_and\_turn}. At turn 24, the combined
\texttt{orientation+target\_loss+zero\_progress} context selects
\texttt{reacquire\_and\_center}. These general strategies are specialized
into the task-specific instruction:
\begin{quote}
\small
Look down, center the crosshair on a sand block, and continuously attack it.
\end{quote}
The resulting harness completes the task in 29 decision turns, compared with
the 100-turn failure of Auto-K. Following success,
\texttt{backup\_and\_turn} and \texttt{reacquire\_and\_center} receive
delayed credits of \(0.8\) and \(1.0\), respectively, allowing the experience
to influence probing on subsequent tasks.

The three settings can therefore be summarized as
\begin{equation}
\underbrace{\text{Bare}}_{\substack{\text{frozen execution}\\
\text{no recovery harness}}}
\;\longrightarrow\;
\underbrace{\text{Auto-K}}_{\substack{\text{knowledge search}\\
\text{fixed probe selection}}}
\;\longrightarrow\;
\underbrace{\text{\method{}}}_{\substack{\text{adaptive action and knowledge probes}\\
\text{closed-loop feedback}}}.
\end{equation}
The key distinction is not whether probe actions are present, but whether
their outcomes are used to test, specialize, and update reusable recovery
knowledge. Because Auto-K and \method{} also acquire different knowledge and
task-specific prompts, this example illustrates their system-level behavior
rather than isolating the causal effect of the probing policy alone.

\section{Theoretical Derivation of Knowledge--Probing Co-Evolution}
\label{app:tthe-theory}

This appendix develops the mathematical rationale for the probing and
validation rules in \method{}. We derive the immediate utility and delayed credit,
characterize the transfer-regularized value estimator, motivate the
UCB-inspired exploration term, and analyze paired harness validation. The
estimation results are conditional on their stated assumptions and describe
individual components of the evolving harness.

\subsection{Context-Conditioned Probe Selection}
\label{app:tthe-context}

Consider either the action-probing policy or the knowledge-probing policy.
Let $x$ denote its task and failure context, and let
$v\in\mathcal V(x)$ denote an available action probe or recovery strategy.
Both policies use the following calculations while maintaining separate
statistics. For a fixed context--candidate pair, define
\begin{equation}
N=N_{x,v},\qquad
Q=Q_{x,v}=\sum_{i=1}^{N}q_i,\qquad
N_x=\sum_{v'\in\mathcal V(x)}N_{x,v'},
\label{eq:tthe-statistics}
\end{equation}
where $q_i$ is the utility assigned to a selection. When $N>0$, the
local empirical mean is $\widehat q_{x,v}=Q/N$. The cross-context mean
$\bar q_v$ provides a shared reference value for candidate $v$.

The local estimation target is
\begin{equation}
\mu_{x,v}=\mathbb E[q\mid x,v].
\label{eq:tthe-local-value}
\end{equation}
This quantity measures expected credited utility under the current execution
conditions. For the statistical results below, we consider a fixed stage with
stationary utility distributions and completed feedback. The operational
system may update its harness and knowledge bank between stages.

\subsection{Immediate Utility as Scalarized Feedback}
\label{app:tthe-immediate}

A selected probe produces task-progress change $\Delta p_t$, environment
reward $r_t$, observation novelty $n_t$, interaction cost $C_t$, and a
failure indicator $F_t=\mathbf 1[\mathrm{fail}_t]$. A constrained local
objective that motivates their combination is
\begin{align}
\max_{\pi}\quad&
\mathbb E_{\pi}[w_p\Delta p_t+w_r r_t+w_n n_t],
\label{eq:tthe-constrained-objective}\\
\text{subject to}\quad&
\mathbb E_{\pi}[C_t]\le C_{\max},\qquad
\mathbb E_{\pi}[F_t]\le F_{\max}.
\nonumber
\end{align}
For nonnegative multipliers $\eta_c$ and $\eta_f$, the Lagrangian is
\begin{align}
\mathcal L(\pi,\eta_c,\eta_f)
={}&
\mathbb E_{\pi}\!\left[
w_p\Delta p_t+w_r r_t+w_n n_t-\eta_cC_t-\eta_fF_t
\right]
\nonumber\\
&+\eta_cC_{\max}+\eta_fF_{\max}.
\label{eq:tthe-lagrangian}
\end{align}
The final two terms are independent of the selected probe. Writing a fixed
per-selection charge as $c_p=\eta_cC_t$ and setting $w_f=\eta_f$ gives
\begin{equation}
q_t^{\mathrm{imm}}
=
w_p\Delta p_t+w_r r_t+w_n n_t
-c_p-w_f\mathbf 1[\mathrm{fail}_t].
\label{eq:tthe-immediate-utility}
\end{equation}
This recovers the immediate utility used in the main text. The coefficients
specify the relative value of progress, reward, novelty, interaction cost,
and failed execution. \method{} uses fixed coefficients rather than optimizing
the dual variables in Equation~\ref{eq:tthe-lagrangian}. Their numerical
values are design hyperparameters reported in the experimental setup.

For action probing, the novelty term rewards newly revealed interaction
evidence. Knowledge probing sets $w_n=0$ and evaluates the progress and
reward following application of the selected recovery strategy. This
scalarization supplies a common update signal for heterogeneous outcomes;
novelty serves as an operational proxy for informative interaction.

\subsection{Delayed Success Credit and Statistical Updates}
\label{app:tthe-delayed}

Let
\begin{equation}
Y=R_{\mathrm{succ}}\mathbf 1[\mathrm{Succ}(\tau,g)]
\label{eq:tthe-terminal-bonus}
\end{equation}
be the terminal-success bonus. For a selection at timestep $t$, let $d_t$
denote its nonnegative distance to terminal feedback on the credit-assignment
clock. Define the terminal contribution at distance $d$ recursively by
\begin{equation}
G(0)=Y,\qquad G(d+1)=\gamma G(d),
\qquad 0<\gamma\le1.
\label{eq:tthe-credit-recursion}
\end{equation}
Repeated substitution yields $G(d)=\gamma^dY$, and hence
\begin{equation}
q_t^{\mathrm{delayed}}
=
R_{\mathrm{succ}}\gamma^{d_t}
\mathbf 1[\mathrm{Succ}(\tau,g)].
\label{eq:tthe-delayed-credit}
\end{equation}
Equivalently, when selection and termination are indexed by the same
discounting clock, a terminal-only reward sequence satisfies
\begin{equation}
b_u=0\ \ (u<T),\qquad b_T=Y,\qquad
\sum_{u=t}^{T}\gamma^{u-t}b_u=\gamma^{T-t}Y.
\label{eq:tthe-terminal-return}
\end{equation}
Under this indexing convention, $d_t=T-t$.

The complete credited utility of a selection is
\begin{equation}
q_t=q_t^{\mathrm{imm}}+q_t^{\mathrm{delayed}}.
\label{eq:tthe-complete-utility}
\end{equation}
It combines local feedback with the discounted terminal outcome.
At selection time, the statistics are updated as
\begin{equation}
N_{x_t,v_t}\leftarrow N_{x_t,v_t}+1,\qquad
Q_{x_t,v_t}\leftarrow Q_{x_t,v_t}+q_t^{\mathrm{imm}}.
\label{eq:tthe-online-update}
\end{equation}
At episode completion, each recorded selection receives the additional update
\begin{equation}
Q_{x_t,v_t}\leftarrow
Q_{x_t,v_t}+q_t^{\mathrm{delayed}}.
\label{eq:tthe-terminal-update}
\end{equation}
The count is incremented only once because both feedback terms belong to the
same selection. Cross-context aggregates incorporate the same updates.

For repeated selections of a given $(x,v)$ within an episode, define the
terminal eligibility
\begin{equation}
e_{x,v}
=
\sum_{j:(x_j,v_j)=(x,v)}\gamma^{d_j}.
\label{eq:tthe-eligibility}
\end{equation}
The total terminal update is then $Q_{x,v}\leftarrow Q_{x,v}+Ye_{x,v}$,
which is algebraically identical to summing the individual delayed bonuses.

Delayed credit records association with successful execution. It does not
identify the causal contribution of an individual intervention. Persistent
updates occur during learning rollouts; paired, holdout, and final frozen
evaluations disable writeback.

\subsection{Transfer-Regularized Value Estimation}
\label{app:tthe-estimation}

For sparse contexts, the local mean $Q/N$ can be noisy and is undefined
when $N=0$. We estimate local utility by regularizing toward the current
cross-context value $\bar q_v$:
\begin{equation}
L(\mu)
=
\frac12\sum_{i=1}^{N}(q_i-\mu)^2
+
\frac{\lambda}{2}(\mu-\bar q_v)^2,
\qquad \lambda>0.
\label{eq:tthe-regularized-objective}
\end{equation}
The first term fits local observations, and the second incorporates
experience accumulated across contexts.

\paragraph{Proposition 1: regularized value estimate.}
Holding $\bar q_v$ fixed during minimization, the unique minimizer of
Equation~\ref{eq:tthe-regularized-objective} is
\begin{equation}
\widehat\mu_{x,v}
=
\frac{Q_{x,v}+\lambda\bar q_v}{N_{x,v}+\lambda}.
\label{eq:tthe-regularized-estimate}
\end{equation}

\paragraph{Proof.}
Differentiating the objective gives
\begin{align}
\frac{\partial L}{\partial\mu}
&=
\sum_{i=1}^{N}(\mu-q_i)+\lambda(\mu-\bar q_v)
\nonumber\\
&=(N+\lambda)\mu-Q-\lambda\bar q_v.
\end{align}
The stationary condition therefore implies
\begin{equation}
(N+\lambda)\widehat\mu_{x,v}=Q+\lambda\bar q_v.
\end{equation}
Since $\partial^2L/\partial\mu^2=N+\lambda>0$, the stationary point is the
unique global minimizer. \hfill$\square$

For $N>0$, this estimate is the convex combination
\begin{equation}
\widehat\mu_{x,v}
=
\frac{N}{N+\lambda}\widehat q_{x,v}
+
\frac{\lambda}{N+\lambda}\bar q_v.
\label{eq:tthe-shrinkage-form}
\end{equation}
Thus $\lambda$ is the effective weight of transferred experience. At $N=0$,
the estimate equals the reference value $\bar q_v$. As local observations
accumulate with fixed $\lambda$, their weight approaches one.

\paragraph{Gaussian interpretation.}
For an idealized Gaussian likelihood
$q_i\mid\mu\sim\mathcal N(\mu,\sigma^2)$ and an independently specified
prior $\mu\sim\mathcal N(\bar q_v,\tau^2)$, the negative log posterior is,
up to an additive constant,
\begin{equation}
\frac{1}{2\sigma^2}\sum_i(q_i-\mu)^2
+
\frac{1}{2\tau^2}(\mu-\bar q_v)^2.
\end{equation}
Multiplication by $\sigma^2$ recovers
Equation~\ref{eq:tthe-regularized-objective} with
$\lambda=\sigma^2/\tau^2$. In \method{}, the cross-context value is data-dependent,
so the regularized optimization interpretation applies directly; the
independent-prior interpretation describes an idealized statistical model.

\paragraph{Proposition 2: bias--variance tradeoff of transfer.}
Suppose $N>0$ is fixed, $q_i$ are independent with mean $\mu$ and variance
$\sigma^2$, and the reference value $m$ is fixed independently of these
samples. Let $d=m-\mu$. Then
\begin{equation}
\operatorname{MSE}(\widehat\mu)
=
\frac{N\sigma^2+\lambda^2d^2}{(N+\lambda)^2}.
\label{eq:tthe-mse}
\end{equation}
The regularized estimator has smaller MSE than the local sample mean
if and only if
\begin{equation}
d^2<\sigma^2\left(\frac{2}{\lambda}+\frac{1}{N}\right).
\label{eq:tthe-transfer-condition}
\end{equation}

\paragraph{Proof.}
Using $\widehat\mu=(\sum_iq_i+\lambda m)/(N+\lambda)$ gives
\begin{align}
\mathbb E[\widehat\mu]-\mu
&=\frac{\lambda(m-\mu)}{N+\lambda},
\nonumber\\
\operatorname{Var}(\widehat\mu)
&=\frac{N\sigma^2}{(N+\lambda)^2}.
\end{align}
Adding squared bias and variance proves Equation~\ref{eq:tthe-mse}.
The unregularized sample mean has MSE $\sigma^2/N$. Consequently,
\begin{align}
\operatorname{MSE}(\widehat\mu)<\frac{\sigma^2}{N}
&\iff
N\sigma^2+\lambda^2d^2
<
\frac{\sigma^2}{N}(N+\lambda)^2
\nonumber\\
&\iff
N\lambda^2d^2
<
2N\lambda\sigma^2+\lambda^2\sigma^2,
\end{align}
which gives Equation~\ref{eq:tthe-transfer-condition}.
\hfill$\square$

The proposition explains how related contexts can improve sparse local
estimation: variance reduction outweighs transfer bias when the reference
value is sufficiently close to the local mean. Its assumptions concern a
fixed reference and sample count; a changing, data-dependent reference
requires additional analysis.

\subsection{Derivation of the UCB-Inspired Selection Score}
\label{app:tthe-ucb}

The exploration term follows the optimism principle underlying UCB \citep{auer2002finite}, with decisions conditioned on task and failure context.
The context-conditioned formulation is related to contextual bandits \citep{li2010contextual}.

\paragraph{Concentration of the local mean.}
Assume that completed utilities for a fixed $(x,v)$ are independent,
stationary, and bounded in $[a,b]$, with width $B=b-a>0$.
For $N>0$, Hoeffding's inequality gives
\begin{equation}
\Pr\!\left(
|\widehat q_{x,v}-\mu_{x,v}|\ge\epsilon
\right)
\le
2\exp\!\left(-\frac{2N\epsilon^2}{B^2}\right).
\label{eq:tthe-hoeffding}
\end{equation}
Setting the right-hand side equal to $\delta\in(0,1)$ yields
\begin{align}
-\frac{2N\epsilon^2}{B^2}
&=\log(\delta/2),
\nonumber\\
r_N(\delta)
&=B\sqrt{\frac{\log(2/\delta)}{2N}}.
\label{eq:tthe-confidence-radius}
\end{align}
Thus the local mean lies within $r_N(\delta)$ of its expectation with
probability at least $1-\delta$.

For a fixed finite set of $M$ candidates, a simultaneous version follows
by allocating
\begin{equation}
\delta_{v,n}=\frac{6\delta}{\pi^2Mn^2},
\qquad n\ge1.
\end{equation}
Because $\sum_{v=1}^{M}\sum_{n=1}^{\infty}\delta_{v,n}=\delta$,
a union bound gives, simultaneously for all candidates and sample counts,
\begin{equation}
|\widehat q_{x,v,n}-\mu_{x,v}|
\le
B\sqrt{\frac{\log(\pi^2Mn^2/(3\delta))}{2n}}
\label{eq:tthe-uniform-bound}
\end{equation}
with probability at least $1-\delta$. Under these stationary sampling
assumptions, the bound can be evaluated at an adaptively selected count.

\paragraph{Practical exploration bonus.}
Taking a confidence schedule
$\delta_x=(1+N_x)^{-\alpha}$, with $\alpha>0$ and $N_x\ge1$, gives
\begin{equation}
r_N(\delta_x)
=
B\sqrt{\frac{\log 2+\alpha\log(1+N_x)}{2N}},
\end{equation}
which motivates the dependence
$\sqrt{\log(1+N_x)/N}$. \method{} adopts the smoothed bonus
\begin{equation}
b(x,v)
=
\beta\sqrt{\frac{\log(1+N_x)}{N_{x,v}+1}},
\qquad \beta\ge0.
\label{eq:tthe-exploration-bonus}
\end{equation}
The additive constants make the expression defined at zero visits.
The coefficient $\beta$ controls the strength of exploration.

Combining Equation~\ref{eq:tthe-regularized-estimate} with
Equation~\ref{eq:tthe-exploration-bonus} gives
\begin{equation}
\operatorname{UCB}(v\mid x)
=
\frac{Q_{x,v}+\lambda\bar q_v}{N_{x,v}+\lambda}
+
\beta
\sqrt{
\frac{\log\!\left(1+\sum_{v'}N_{x,v'}\right)}
{N_{x,v}+1}
}.
\label{eq:tthe-full-selection-score}
\end{equation}
The policy selects
\begin{equation}
v_t\in
\operatorname*{arg\,max}_{v\in\mathcal V(x)}
\operatorname{UCB}(v\mid x).
\end{equation}
The first term estimates utility using local and cross-context experience;
the second favors less frequently selected candidates. At zero total visits,
the exploration term is zero and selection is determined by reference values
and tie-breaking.

\paragraph{Confidence interpretation with transfer.}
The regularized estimate satisfies the identity
\begin{align}
\mu_{x,v}-\widehat\mu_{x,v}
={}&
\frac{N}{N+\lambda}
(\mu_{x,v}-\widehat q_{x,v})
\nonumber\\
&+
\frac{\lambda}{N+\lambda}
(\mu_{x,v}-\bar q_v).
\end{align}
On the concentration event in
Equation~\ref{eq:tthe-confidence-radius}, the triangle inequality gives
\begin{equation}
|\mu_{x,v}-\widehat\mu_{x,v}|
\le
\frac{N}{N+\lambda}r_N(\delta)
+
\frac{\lambda}{N+\lambda}
|\bar q_v-\mu_{x,v}|.
\label{eq:tthe-transfer-error}
\end{equation}
This inequality remains valid for a data-dependent reference value.
If $|\bar q_v-\mu_{x,v}|\le D_{x,v}$ is additionally known, a corresponding
upper bound is
\begin{equation}
U^{\mathrm{cert}}_{x,v}
=
\widehat\mu_{x,v}
+
\frac{N}{N+\lambda}r_N(\delta)
+
\frac{\lambda}{N+\lambda}D_{x,v}.
\label{eq:tthe-certified-bound}
\end{equation}
Equation~\ref{eq:tthe-full-selection-score} uses a smoothed exploration
bonus without explicitly estimating this transfer-bias term. We therefore
describe the implemented rule as UCB-inspired. Its design is motivated by
concentration, while the conditional bound above identifies the additional
quantity required for a calibrated confidence interpretation.

\subsection{Paired Harness Validation}
\label{app:tthe-validation}

Fix an incumbent harness $H_k$ and a candidate $\widetilde H_k$.
For an evaluation condition $(g,s)$, define
\begin{equation}
D(g,s)
=
J(\widetilde H_k;g,s)-J(H_k;g,s),
\qquad
\Delta J_k=\mathbb E[D(g,s)].
\label{eq:tthe-paired-difference}
\end{equation}
Here $s$ indexes evaluation randomization, and both harnesses use the same
task configuration, score, and initial external-state snapshot.
For $m=|V_k|$ paired conditions, the empirical improvement is
\begin{equation}
\widehat{\Delta J}_k
=
\frac{1}{m}
\sum_{(g,s)\in V_k}
\left[
J(\widetilde H_k;g,s)-J(H_k;g,s)
\right].
\label{eq:tthe-paired-mean}
\end{equation}
For a candidate fixed independently of identically distributed evaluation
draws, linearity of expectation gives
\begin{equation}
\mathbb E[\widehat{\Delta J}_k]
=
\frac1m\sum_{i=1}^{m}\mathbb E[D_i]
=
\Delta J_k.
\end{equation}

The variance of a paired difference is
\begin{align}
\operatorname{Var}(D)
={}&
\operatorname{Var}(J_{\mathrm{cand}})
+
\operatorname{Var}(J_{\mathrm{inc}})
\nonumber\\
&-
2\operatorname{Cov}(J_{\mathrm{cand}},J_{\mathrm{inc}}).
\label{eq:tthe-pair-variance}
\end{align}
Positive covariance under shared task conditions reduces the variance
relative to independent evaluation. Isolated external-state copies and
disabled writeback prevent one condition from changing the persistent
state used by the other.

\paragraph{Acceptance as an empirical decision.}
Let $\widehat J(H)$ denote the mean score on the paired conditions.
Charging an improvement margin $\epsilon_{\mathrm{pair}}$ for replacing
the incumbent gives
\begin{equation}
H_{k+1}\in
\operatorname*{arg\,max}_{H'\in\{H_k,\widetilde H_k\}}
\left[
\widehat J(H')
-
\epsilon_{\mathrm{pair}}
\mathbf 1[H'=\widetilde H_k]
\right].
\end{equation}
With ties resolved in favor of the candidate, the candidate is selected
exactly when
\begin{equation}
\widehat J(\widetilde H_k)-\epsilon_{\mathrm{pair}}
\ge\widehat J(H_k),
\end{equation}
or equivalently
$\widehat{\Delta J}_k\ge\epsilon_{\mathrm{pair}}$. Thus
\begin{equation}
\phi_{k+1}
=
\begin{cases}
\widetilde\phi_k,
&\widehat{\Delta J}_k\ge\epsilon_{\mathrm{pair}},\\
\phi_k,
&\text{otherwise}.
\end{cases}
\label{eq:tthe-acceptance-rule}
\end{equation}

\paragraph{Conditional statistical interpretation.}
Suppose $J\in[A_J,B_J]$ with width $W_J=B_J-A_J$, and the $m$ paired
differences are independent for a candidate fixed before evaluation.
Then $D_i\in[-W_J,W_J]$, and one-sided Hoeffding gives
\begin{equation}
\Pr\!\left(
\widehat{\Delta J}_k-\Delta J_k\ge\epsilon
\right)
\le
\exp\!\left(-\frac{m\epsilon^2}{2W_J^2}\right).
\end{equation}
Consequently, with probability at least $1-\delta$,
\begin{equation}
\Delta J_k
\ge
\widehat{\Delta J}_k
-
W_J\sqrt{\frac{2\log(1/\delta)}{m}}.
\label{eq:tthe-paired-lower-bound}
\end{equation}
A positive lower bound would establish expected improvement under these
assumptions. The implemented acceptance margin is an empirical criterion,
and adaptive reuse of evaluation data requires additional statistical
control. Holdout assessment provides separate evidence after acceptance
and does not alter the original paired decision.

\subsection{Evidence-Conditioned Knowledge Selection}
\label{app:tthe-evidence}

For a fixed harness state, let $a$ be an action probe, $y$ the resulting
evidence, and $z$ the selected recovery knowledge. The two-stage selection
process can be written as
\begin{equation}
\pi(a,z\mid x,\mathcal K)
=
\pi_a(a\mid x)
\int
p(y\mid x,a)\,
\pi_z(z\mid x,a,y,\mathcal K)\,\mathrm dy.
\label{eq:tthe-joint-policy}
\end{equation}
For discrete evidence, the integral is replaced by a sum.
Action probing changes the evidence available to knowledge selection.
Execution feedback updates both selection policies, while validated
recoveries expand $\mathcal K$.

\paragraph{Proposition 3: value of diagnostic information.}
Consider an idealized diagnostic probe that has no execution cost and
does not change the physical state. Let $F$ denote the hidden failure
cause, let $\mathcal Z$ be a fixed finite set of recovery strategies,
and let $u(z,F)$ be their integrable utility. Define
\begin{align}
V_0(x)
&=
\max_{z\in\mathcal Z}
\mathbb E[u(z,F)\mid x],
\nonumber\\
V_1(x)
&=
\mathbb E_y\!\left[
\max_{z\in\mathcal Z}
\mathbb E[u(z,F)\mid x,y]
\mid x
\right].
\end{align}
Then $V_1(x)\ge V_0(x)$.

\paragraph{Proof.}
Let $z^\star$ maximize the objective defining $V_0(x)$. For every observed
$y$, the maximizing strategy performs at least as well in conditional
expectation as retaining $z^\star$. Therefore,
\begin{align}
V_1(x)
&\ge
\mathbb E_y\!\left[
\mathbb E[u(z^\star,F)\mid x,y]
\mid x
\right]
\nonumber\\
&=
\mathbb E[u(z^\star,F)\mid x]
=
V_0(x),
\end{align}
where the equality follows from the law of iterated expectations.
\hfill$\square$

The proposition explains the role of diagnostic evidence in knowledge
selection. Real probes incur costs, may change the state, and are followed
by approximate decisions, so their realized value is evaluated through
interaction. In \method{}, this evaluation also changes future probe preferences
and the available recovery knowledge, providing the feedback coupling
underlying knowledge--probing co-evolution.

The regularized estimator and credit updates follow exactly from their
specified objectives. The concentration and transfer results apply under
the stated local sampling assumptions. Harness evolution, changing recovery
policies, and pending terminal feedback can violate stationarity or complete
observation. Accordingly, the analysis explains the components of \method{}
without asserting a global regret or monotonic-improvement guarantee for
the full evolving system.

\section{Computational Cost and Interaction Budget}
\label{sec:computational_cost}

Our evaluation uses a two-stage trained VLA:
\[
\text{Qwen3-VL-8B}
\xrightarrow{\text{cold-start SFT}}
C_{16200}
\xrightarrow{\text{GRPO}}
\pi_{400}.
\]
The resulting GRPO checkpoint $\pi_{400}$ is then frozen and shared by
Bare, Fixed, Auto-K, and \method{}. Thus, no method-specific gradient update
is performed during knowledge search, adaptive probing, or evaluation.
Nevertheless, both the upstream actor training and the subsequent
closed-loop sandbox interaction incur substantial computational cost.

\paragraph{Cold-start supervised fine-tuning.}
The cold-start stage performs full-parameter supervised fine-tuning of the
Qwen3-VL-8B actor using 16 GPUs. The GRPO stage is initialized from the
cold-start checkpoint at step 16,200. At this checkpoint, cold-start training
has completed 2.72 epochs and consumed 121.3 hours of wall-clock time,
corresponding to approximately 1,940.3 GPU-hours.

\paragraph{GRPO post-training.}
Starting from the cold-start checkpoint $C_{16200}$, the actor is further
post-trained with GRPO using 32 H800 GPUs: eight GPUs for generation and
24 GPUs for actor optimization, with actor data parallelism 12 and tensor
parallelism 2. The final actor used in all experiments is checkpoint 400.
The training underwent several infrastructure restarts. Summing the active
job intervals recorded by TensorBoard up to checkpoint 400 gives at least
379.8 hours of cumulative 32-GPU job time, corresponding to at least
12,155 GPU-hours. This is a lower bound because launches that failed before
writing training events are not observable in the retained logs.

\paragraph{Harness evolution and evaluation.}
After GRPO, the VLA is frozen. Harness evolution still requires repeated VLA
inference and interaction with live Minecraft sandboxes. The major harness
experiments use eight H800 GPUs, with one worker bound to each GPU. For these
experiments, \emph{summed worker time} aggregates elapsed time across
single-GPU workers and therefore represents a serial-equivalent interaction
cost rather than eight-GPU wall-clock time.

\begin{table*}[t]
\centering
\caption{
Computational cost of actor training, harness evolution, and evaluation.
GPU-hours are not normalized across accelerator types. Values preceded by
$\geq$ are conservative lower bounds reconstructed from retained artifacts.
}
\label{tab:computational_cost}
\resizebox{\textwidth}{!}{
\begin{tabular}{l|c|c|r|l|r}
\hline
Stage
& Parallel GPUs
& Training/interaction budget
& Environment steps
& Time accounting
& GPU-hours \\
\hline
Cold-start SFT
& 16
& 16,200 updates
& --
& 121.3 h wall-clock
& 1,940.3 \\

GRPO post-training
& 32
& 400 updates
& --
& $\geq$379.8 h cumulative active job time
& $\geq$12,155.0 \\
\hline

Knowledge search and co-evolution
& 8
& 1,861 episodes
& 175,013
& $\geq$547.5 h summed worker time
& $\geq$547.5 \\

Knowledge--probing ablation
& 8
& 440 episodes
& 41,777
& 124.6 h summed worker time
& $\geq$124.6 \\

Frozen transfer evaluation
& 8
& 180 episodes
& 16,709
& 6.94 h job wall-clock
& 55.5 \\
\hline

\multicolumn{5}{r|}{Total recorded allocation}
& $\geq$14,822.9 \\
\hline
\end{tabular}
}
\end{table*}

Across the post-training harness experiments, the retained artifacts contain
2,481 sandbox episodes, 233,499 environment steps, and 19,310 executed probe
actions. Knowledge search additionally produces 831 recorded proposer
responses and requires at least 1,654 external request attempts, with
approximately 5.76 million provider-reported tokens. The request and token
counts are also lower bounds because failed requests do not always return
usage metadata.

The distinction between worker time and job wall-clock is important. For
example, frozen transfer takes 6.94 hours on eight H800 GPUs, corresponding
to 55.5 allocated GPU-hours, while its completed episodes account for
48.8 summed interaction hours. The remaining allocation is explained by
model loading, sandbox initialization, synchronization, and worker
imbalance. Similarly, the reported 547.5 hours for knowledge search is
already summed across single-GPU workers and should not be multiplied by
eight.

Overall, the complete pipeline requires at least 14.8K recorded GPU-hours,
of which GRPO post-training is the dominant component. Harness evolution is
parameter-update-free but not computation-free: its cost is dominated by
repeated VLA inference, sandbox interaction, candidate validation, and
external proposer calls.

\section{AI use statement}

We used an LLM for language polishing and for checking the consistency and
completeness of mathematical notation. We also used Gemini-2.5-Flash through an API
as the vision-language researcher and harness proposer in our experiments,
as described in the methodology. Its proposals were accepted after
sandbox execution and paired validation.

\section{Ethics Statement}

This work studies test-time harness evolution for a frozen
vision--language--action model in the simulated Minecraft environment. It
does not involve human subjects, personally identifiable information, or
real-world deployment. All evaluations are conducted in controlled virtual
environments. The method introduces additional computation through environment
interaction, VLA inference, and multimodal researcher calls; we acknowledge
the associated computational and energy costs and report the main resource
requirements in the appendix. As with other adaptive agent systems, deployment
in real-world environments may introduce additional safety and reliability
considerations that are outside the scope of this work.

\section{Reproducibility Statement}

The paper and
appendix report the VLA training configuration, evaluation protocols,
probing and credit-assignment hyperparameters, harness search procedure,
paired acceptance rule, low-level action space, and computational resources.
We also provide the task configurations and scripts required to reproduce the
reported experiments where permitted by the corresponding environments and
model APIs. External model calls use the model identifiers and decoding
settings specified in the experimental setup.

\section{Computational Resources}
SFT and GRPO training were conducted on NVIDIA H800 GPUs with
141\,GB of memory per GPU. Policy evaluation, harness evolution, probing,
and sandbox validation were performed on NVIDIA H800 GPUs with 80\,GB of
memory per GPU.

\end{document}